\documentclass[conference]{IEEEtran}

\pdfoutput=1

\usepackage{graphicx}
\usepackage{epsfig} 
\usepackage{mathptmx}
\usepackage{amsmath} 

\usepackage{amssymb}  
\usepackage{amsthm}
\usepackage{wasysym}
\usepackage[mathcal]{euscript}
\usepackage{mathrsfs}
\usepackage{bm}
\usepackage{bbm}
\usepackage{color}
\usepackage{lipsum}
\usepackage[ruled,vlined,linesnumbered]{algorithm2e}
\usepackage[colorlinks=true,linkcolor=black,anchorcolor=black,citecolor=black,filecolor=black,menucolor=black,runcolor=black,urlcolor=cyan]{hyperref}
\usepackage{etoolbox}
\usepackage[table,xcdraw,dvipsnames]{xcolor}
\usepackage{multirow}
\usepackage{mathtools}
\usepackage{algpseudocode}
\usepackage{xspace}

\hypersetup{
    colorlinks=true,
    linkcolor=blue,
    filecolor=magenta,      
    urlcolor=cyan,
    pdftitle={Overleaf Example},
    pdfpagemode=FullScreen,
    }

\usepackage{outlines}
\let\labelindent\relax
\usepackage{enumitem}
\setenumerate[1]{label=\Roman*.}
\setenumerate[2]{label=\Alph*.}
\setenumerate[3]{label=\arabic*.}
\setenumerate[4]{label=\roman*.}

\usepackage{pifont}

\usepackage{booktabs}

\makeatletter
\patchcmd{\@makecaption}
  {\scshape}
  {}
  {}
  {}
\makeatother
\usepackage[dvipsnames]{xcolor}
\usepackage{changes}

\newcommand{\stkout}[1]{{\color{Plum}\ifmmode\text{\sout{\ensuremath{#1}}}\else\sout{#1}\fi}}

\newcommand{\shrug}{\texttt{\raisebox{0.75em}{\char`\_}\char`\\\char`\_\kern-0.5ex(\kern-0.25ex\raisebox{0.25ex}{\rotatebox{45}{\raisebox{-.75ex}"\kern-1.5ex\rotatebox{-90})}}\kern-0.5ex)\kern-0.5ex\char`\_/\raisebox{0.75em}{\char`\_}}}

\providecommand{\methodname}{\text{SPARROWS}\xspace}
\providecommand{\armtd}{\text{ARMTD}\xspace}
\providecommand{\mpot}{\text{MPOT}\xspace}
\providecommand{\chomp}{\text{CHOMP}\xspace}

\providecommand{\trajopt}{\text{TrajOpt}\xspace}

\newcommand{\abs}[1]{\left\vert#1\right\vert}

\newtheorem{defn}{Definition}

\newtheorem{lem}[defn]{Lemma}

\newtheorem{assum}[defn]{Assumption}

\newtheorem{thm}[defn]{Theorem}

\newcommand{\regtext}[1]{\mathrm{\textnormal{#1}}}

\newcommand{\defemph}[1]{\emph{#1}}
\newcommand{\ts}[1]{\textsuperscript{#1}}
\newcommand{\nd}{n_d}

\providecommand{\R}{\ensuremath \mathbb{R}}
\providecommand{\N}{\ensuremath \mathbb{N}}

\providecommand{\nan}{\texttt{NaN}\xspace}

\providecommand{\tplan}{t_p}

\providecommand{\opt}{\texttt{(Opt)}}
\providecommand{\optref}{\hyperref[eq:pzoptcost]{\opt{}}}

\providecommand{\world}{W}
\providecommand{\workspace}{W_s}

\newcommand{\nObs}{n}
\newcommand{\NObs}{ N_{\mathscr{O}} }
\newcommand{\obsset}{\mathscr{O}}

\newcommand{\nq}{n_q}
\newcommand{\ns}{n_s}
\newcommand{\nt}{n_t}

\newcommand{\Nq}{ N_q }

\newcommand{\Nt}{ N_t }

\providecommand{\FO}{\regtext{\small{FO}}}

\newcommand{\setop}[1]{{\mathrm{\textnormal{\texttt{#1}}}}}
\newcommand{\numop}[1]{{\mathrm{\textnormal{\texttt{#1}}}}}

\newcommand{\qlim}{q_{j,\regtext{lim}}}
\newcommand{\dqlim}{\dot{q}_{j,\regtext{lim}}}

\newcommand{\timestep}{\Delta t}

\newcommand{\conv}{co}

\providecommand{\tfin}{t_\text{f}}

\newcommand{\kjvar}{x_{k_j}}

\newcommand{\Kj}{K_j}

\newcommand{\dist}{d}
\newcommand{\sdf}{s_d}
\newcommand{\SDF}{\texttt{SDF}}
\newcommand{\proj}{\phi}

\newcommand{\zeros}{\textit{0}}

\newcommand{\pow}[1]{\mathcal{P}\!\left(#1\right)}

\providecommand{\sjo}{\mathcal{SJO}}

\providecommand{\sfo}{\mathcal{SFO}}

\newcommand{\SJO}{{Spherical Joint Occupancy}\xspace}

\newcommand{\SFO}{{Spherical Forward Occupancy}\xspace}

\newcommand{\iv}[1]{[ #1 ]}

\newcommand{\pz}[1]{\mathbf{#1}}

\newcommand{\PZ}[1]{\mathcal{PZ}\left(#1\right)}

\newcommand{\pzi}[1]{\pz{ #1 }(\pz{T_i};\pz{K})}

\newcommand{\pzki}[1]{\pz{ #1 }(\pz{T_i};k)}

\newcommand{\pzqi}{\pzi{q}}
\newcommand{\pzqli}{\pzi{q_l}}

\newcommand{\pzqki}{\pzki{q}}

\newcommand{\pzqji}{\pzi{q_j}}
\newcommand{\pzqdji}{\pzi{\dot{q}_j}}

\newcommand{\pzqjki}{\pzki{q_j}}
\newcommand{\pzqdjki}{\pzki{\dot{q}_j}}

\newcommand{\pzFKjki}{\pz{FK_j}(\pzqki)}

\newcommand{\pzFKjKi}{\pz{FK_j}(\pzqi)}
\newcommand{\pzFOjKi}{\pz{FO_j}(\pzqi)}

\newcommand{\pzv}{x}

\newcommand{\pzn}{{n_g}}

\newcommand{\pzgi}{g_i}

\newcommand{\pzei}{\alpha_i}

\newcommand{\tvari}{x_{t_{i}}}

\newcommand{\q}{q(t)}

\newcommand{\qstart}{q_{\text{start}}}

\newcommand{\qj}{q_j(t)}
\newcommand{\ql}{q_l(t)}
\newcommand{\qdj}{\dot{q}_{j}(t)}

\newcommand{\qkj}{q_j(t; k)}
\newcommand{\qdkj}{\dot{q}_{j}(t; k)}

\newcommand{\homtrans}{H}

\newcommand{\FK}{\regtext{\small{FK}}}

\newcommand{\pzCjiK}{\pz{C_j}(\pz{T_i}; \pz{K})}
\newcommand{\pzCjik}{\pz{C_j}(\pz{T_i}; k)}
\newcommand{\pzCjpik}{\pz{C_{j+1}}(\pz{T_i}; k)}
\newcommand{\pzuj}{\pz{u_j}(\pz{T_i})}

\newcommand{\Sjq}{S_j(\q)}
\newcommand{\Sjqk}{S_j(q(t;k))}
\newcommand{\Sjpq}{S_{j+1}(\q)}

\newcommand{\Sjik}{S_{j}(\pzqki)}

\newcommand{\SjiK}{{S_{j}(\pzi{q})}}

\newcommand{\Sjpik}{S_{j+1}(\pzqki)}

\newcommand{\sjik}{s_{j,i}(k)}
\newcommand{\spjik}{s'_{j,i}(k)}

\newcommand{\Sbarjimk}{{\bar{S}_{j,i,m}(\pzqki)}}

\newcommand{\TCjik}{TC_{j}(\pzqki)}

\providecommand{\sfok}{\mathcal{SFO}(\pzqki)}
\providecommand{\sfojk}{\mathcal{SFO}_j(\pzqki)}

\usepackage[
    style=ieee,
    doi=false,
    isbn=false,
    url=false,
    eprint=false,
    backend=biber,
    natbib=true
    ]{biblatex}
    
\bibliography{references}

\begin{document}

\title{Safe Planning for Articulated Robots Using Reachability-based Obstacle Avoidance With Spheres}
\author{Jonathan Michaux, Adam Li, Qingyi Chen, Che Chen, Bohao Zhang, and Ram Vasudevan}
\thanks{Jonathan Michaux, Adam Li, Qingyi Chen, Che Chen, Bohao Zhang and Ram Vasudevan are with the Department of Robotics, University of Michigan, Ann Arbor, MI 48109. \texttt{\{jmichaux, adamli, chenqy, cctom, jimzhang, ramv\}@umich.edu}.}
\thanks{This work is supported by the Ford Motor Company via the Ford-UM Alliance under award N022977.}


\maketitle
\IEEEpeerreviewmaketitle

\begin{abstract}
Generating safe motion plans in real-time is necessary for the wide-scale deployment of robots in unstructured and human-centric environments. 
These motion plans must be safe to ensure humans are not harmed and nearby objects are not damaged.
However, they must also be generated in real-time to ensure the robot can quickly adapt to changes in the environment.
Many trajectory optimization methods introduce heuristics that trade-off safety and real-time performance, which can lead to potentially unsafe plans.
This paper addresses this challenge by proposing Safe Planning for Articulated Robots Using Reachability-based Obstacle Avoidance With Spheres (\methodname).
\methodname is a receding-horizon trajectory planner that utilizes the combination of a novel reachable set representation and an exact signed distance function to generate provably-safe motion plans. 
At runtime, \methodname uses parameterized trajectories to compute reachable sets composed entirely of spheres that overapproximate the swept volume of the robot's motion.
\methodname then performs trajectory optimization to select a safe trajectory that is guaranteed to be collision-free.
We demonstrate that \methodname' novel reachable set is significantly less conservative than previous approaches.
We also demonstrate that \methodname outperforms a variety of state-of-the-art methods in solving challenging motion planning tasks in cluttered environments.
Code, data, and video demonstrations can be found at \url{https://roahmlab.github.io/sparrows/}.
\end{abstract}

\section{Introduction}\label{sec:intro}
Robot manipulators have the potential to make large, positive impacts on society and improve human lives.
These potential impacts range from replacing humans performing dangerous and difficult tasks in manufacturing and construction to assisting humans with more delicate tasks such as surgery or in-home care.
In each of these settings, the robot must remain safe at all times to prevent harming nearby humans, colliding with obstacles, or damaging high-value objects.
It is also essential that the robot generate motion plans in real-time so as to perform any given task efficiently or to quickly adjust their behavior to react to changes in the environment.

\begin{figure}[t]
    \centering
    \includegraphics[width=0.93\columnwidth]{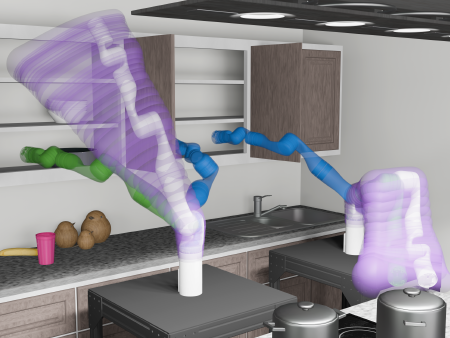}
    \caption{%
    This paper presents \methodname, a method that is capable of generating safe motion plans in dense and cluttered environments for single- and multi-arm robots. 
    Here, both arms have start  and goal configurations shown in blue and green, respectively.
    Prior to planning, \methodname is given full access to polytope overapproximations of obstacles to compute an exact signed distance function of the scene geometry.
    At runtime, \methodname combines the signed distance function with the novel \SFO as obstacle-avoidance constraints for \methodname to generate safe trajectories between the start and goal in a receding horizon manner.
    Each trajectory is selected by solving a nonlinear optimization problem such that an overapproximation (purple) of the swept volume of its entire motion remains collision-free.
    Note in this figure, a small, thin barrier is placed to prevent unwanted collisions between the two arms.}
    \label{fig:network}
    \vspace*{-0.5cm}
\end{figure}

Modern model-based motion planning frameworks typically consist of a high-level planner, a mid-level trajectory planner, and a low-level tracking controller. 
The high-level planner generates a path consisting of a sequence of discrete waypoints between the robot's start and goal configurations. 
The mid-level trajectory planner calculates velocities and accelerations at specific time intervals that move the robot from one waypoint to the next. 
The low-level tracking controller generates control inputs that attempt to minimize deviations from the desired trajectory. 
For example, one may use a high-level sampling-based planner such as a Probabilistic Road Map \cite{kavraki1996probabilistic} to generate discrete waypoints for a trajectory optimization algorithm such as \chomp \cite{Zucker2013chomp} or \trajopt \cite{Schulman2014trajopt}, and track the resulting trajectories with an appropriately designed low-level inverse dynamics controller \cite{Caccavale2020}.
While variations of this framework have been demonstrated to work on various robotic platforms, there are several limitations that prevent wide-scale deployment in the real-world. 
For instance, this approach can become computationally demanding as the complexity of the robot or environment increases, making it less practical for real-time applications. 
Many algorithms also introduce heuristics such as reducing the number of collision checks to achieve real-time performance at the expense of robot safety.  
Furthermore, it is often assumed that the robot's dynamics are fully known, while in reality there can be considerable uncertainty. 
Unfortunately, both of these notions increase the potential for the robot to collide with obstacles.


Reachability-based Trajectory Design (RTD) \cite{kousik2017safe} is a recent example of a hierarchical planning framework that uses a mid-level trajectory planner to enforce safety.
RTD combines zonotope arithmetic \cite{guibas2003zonotopes} with classical recursive robotics algorithms \cite{spong2005textbook} to iteratively compose reachable sets for high dimensional systems such as robot manipulators \cite{holmes2020armtd}.
During planning,  reachable sets are constructed at runtime and used to enforce obstacle avoidance in continuous-time.
In addition to generating collision-free trajectories, RTD has been demonstrated to select trajectories that are guaranteed to be dynamically feasibly even in the presence of uncertainty \cite{michaux2023armour}.
Notably, RTD is able to accomplish certifiably-safe planning in real-time.
However, as we show here, the RTD reachable set formulation for manipulators \cite{holmes2020armtd,michaux2023armour,brei2024waitr} tends to be overly conservative.
This work proposes a novel reachable set formulation composed entirely of spheres that is tighter than that of RTD.
We demonstrate that this novel reachable set formulation enables planning in extremely cluttered environments, whereas RTD fails to find feasible paths, largely due to the size of its reachable sets.

\subsection{Related Work}\label{sec:related_work}
The purpose of a safe motion planning algorithm is to ensure that a robot can move from one configuration to another while avoiding collisions with all obstacles in its environment for all time.
Ideally, one would ensure that the entire swept volume \cite{blackmore_differential_1990, blackmore_analysis_1992, blackmore_analysis_1994, blackmore_sweep-envelope_1997, blackmore_trimming_1999} of a moving robot does not intersect with any objects nor enter any unwanted regions of its workspace.
Although swept volume computation has long been used for collision detection during motion planning \cite{spatial_planningPerez}, computing the exact swept volumes of articulated robots such as serial manipulators is analytically intractable \cite{abdel-malek_swept_2006}.
Instead, many algorithms rely on approximations using CAD models \cite{Campen2010PolygonalBE, Kim2003FastSV, Gaschler2013RobotTA}, occupancy grids, and convex polyhedra.
However, these methods often suffer from high computational costs and are generally not suitable when generating complex robot motions and are therefore most useful when applied while motion planning offline \cite{perrin2012svfootstep}.
Furthermore, approximate swept volume algorithms may be overly conservative \cite{Gaschler2013RobotTA, ekenna2015} and thus limit the efficiency of finding a feasible path in cluttered environments.
To address some of these limitations, \cite{Murray2016RobotMP} used a precomputed swept volume to perform a parallelized collision check at run-time.
However, this approach was not used to generate trajectories in real-time. 

An alternative to computing swept volumes for obstacle avoidance is to model the robot, or the environment, with simple geometric primitives such as spheres \cite{duenser2018manipulation, gaertner2021collisionfree}, ellipsoids \cite{brito2020model}, capsules \cite{dube2013humanoids,khoury2013humanoids}, or convex polygons and perform collision-checking along a given trajectory at discrete time instances.
This is common for state-of-the-art trajectory optimization-based approaches such as \chomp\cite{Zucker2013chomp}, \trajopt \cite{Schulman2014trajopt}, and \mpot \cite{le2023mpot}.
\chomp represents the robot as a collection of discrete spheres and maintains a safety margin with the environment by utilizing a signed distance field.
\trajopt represents the robot and obstacles as the support mapping of convex shapes.
Then the distance and penetration depth between two convex shapes is computed by the Gilbert-Johnson-Keerthi \cite{gtk1988} and Expanding Polytope Algorithms \cite{bergen2001penetration}, respectively.
\mpot represents the robot geometry and obstacles as spheres and implements a collision cost using an occupancy map \cite{le2023mpot}.
Although these approaches have been demonstrated to solve challenging motion planning tasks in real-time, the resulting trajectories cannot be considered safe as collision avoidance is only enforced as a soft penalty in the cost function.

Reachability-based Trajectory Design (RTD) \cite{kousik2017safe} is a recent approach to real-time motion planning that generates provably safe trajectories in a receding-horizon fashion.
At runtime, RTD uses (polynomial) zonotopes \cite{kochdumper2020polyzono} to construct reachable sets that overapproximate all possible robot positions corresponding to a pre-specified continuum of parameterized trajectories.   
RTD then solves a nonlinear optimization problem to select a feasible trajectory such that the swept volume corresponding to that motion is guaranteed to be collision-free.
If a feasible trajectory is not found, RTD executes a braking maneuver that brings the robot safely to a stop.
Importantly, the reachable sets are constructed such that obstacle-avoidance constraints are satisfied in continuous-time.
However, while extensions of RTD have demonstrated real-time, certifiably-safe motion planning for robotic arms \cite{holmes2020armtd, michaux2023armour, brei2024waitr} with seven degrees of freedom, RTD's reachable sets tend to be overly conservative as we demonstrate in this paper.


\subsection{Contributions}
To address the limitations of existing approaches, this paper proposes Safe Planning for Articulated Robots Using Reachability-based Obstacle Avoidance With Spheres (\methodname).
The proposed method combines reachability analysis with sphere-based collision primitives and an exact signed distance function to enable real-time motion planning that is certifiably-safe, yet less conservative than previous methods. 
This paper's contributions are three-fold:
\begin{enumerate}[label={\Roman*.}]
    \item A novel reachable set representation composed of overlapping spheres, called the \SFO ($\sfo$), that overapproximates the robot's reachable set and is differentiable;
    \item An algorithm that computes the exact signed distance between a point and a three dimensional zonotope;
    \item A demonstration that \methodname{} outperforms similar state-of-the-art methods on a set of challenging motion planning tasks.
\end{enumerate}

The remainder of this manuscript is organized as follows:
Section \ref{sec:prelim} summarizes the set representations used throughout the paper and gives a brief overview of signed distance functions;
Section \ref{sec:modeling} describes how the robot arm and environment are modeled;
Section \ref{sec:planning} describes the formulation of the safe motion planning problem using the \SFO and an exact signed distance function;
Section \ref{sec:demonstrations} summarizes the evaluation of the proposed method on a variety of different example problems.


\section{preliminaries}\label{sec:prelim}
This section establishes the notation used throughout the paper and describes the set-based representations and operations used throughout this document.

\subsection{Notation}
Sets and subspaces are typeset using capital letters.
Subscripts are primarily used as an index or to describe a particular coordinate of a vector.
Let $\R$ and $\N$ denote the spaces of real numbers and natural numbers, respectively.
The Minkowski sum between two sets $\mathcal A$ and $\mathcal A'$ is $\mathcal A\oplus \mathcal A' = \{a+a'\mid a\in \mathcal A, ~a'\in \mathcal A'\}$.
Let $\hat{e}_l \in \R^n$ denote the $l$\ts{th} unit vector in the standard orthogonal basis.
Given vectors $\alpha$, let $[\alpha]_i$ denote the $i$-th element of $\alpha$. 
Given $\alpha \in \R^n$ and $\epsilon > 0 $, let $B(\alpha, \epsilon)$ denote the $n$-dimensional closed ball with center $\alpha$ and radius $\epsilon$ under the Euclidean norm.
Given vectors $\alpha,\beta\in\R^n$, let $\alpha \odot \beta$ denote element-wise multiplication. 
Given a set $\Omega \subset \mathbb{R}^{\nd}$, let $\partial \Omega \subset \mathbb{R}^{\nd}$ be its boundary, $\Omega^{c} \subset \mathbb{R}^{\nd}$ denote its complement, and $\conv(\Omega)$ denote its convex hull.

\subsection{Polynomial Zonotopes}
We present a brief overview of the defintions and operations on polynomial zonotopes (PZs) that are used throughout the remainder of this document. 
A thorough introduction to polynomial zonotopes is available in \cite{kochdumper2020polyzono}. 

Given a \defemph{center} $c \in \R^n$, \defemph{dependent generators} $\pzgi \in \R^n$, \defemph{independent generators} $h_j \in \R^n$, and \defemph{exponents} $\pzei \in \N^\pzn_i$ for $i \in \{0 ,\ldots, \pzn\}$, a \defemph{polynomial zonotope} is a set:
\begin{align}\label{eq:pz_definition}
    \hspace*{-0.3cm} \pz{P} 
     = \big\{
                z \in \R^n \, \mid \,
                z = c + \sum_{i=1}^{\pzn} \pzgi \pzv ^{\pzei} + \sum_{j=1}^{n_h} h_j y_j, \, 
                \pzv \in [-1, 1]^\pzn,\\ y_j \in [-1, 1]
            \big\}.
\end{align}
For each $i \in \{1 ,\ldots, \pzn\}$, we refer to $\pzv ^{\pzei}$ as a \defemph{monomial}.
We refer to $\pzv \in [-1,1]^\pzn$ and $y_j \in [-1,1]$ as \defemph{indeterminates} for the dependent and independent generators, respectively. 
Note that one can define a matrix polynomial zonotope by replacing the vectors above with matrices of compatible dimensions.


Throughout this document, we exclusively use bold symbols to denote polynomial zonotopes.
We also introduce the shorthand $\pz{P} = \PZ{c, \pzgi, \pzei, \pzv, h_j, y_j}$ when we need to emphasize the generators and exponents of a polynomial zonotope.
Note that zonotopes are a special subclass of polynomial zonotopes whose dependent generators are all zero \cite{kochdumper2020polyzono}.

\subsection{Polynomial Zonotope Operations}

\begin{table}[t]
    \caption{Summary of polynomial zonotope operations.}
    \centering
    \begin{tabular}{l|c}
        \multicolumn{1}{c|}{Operation} & Computation \\
        \hline
        $\pz{P}_1 \oplus \pz{P}_2$ (Minkowski Sum) \cite[eq. (10)]{michaux2023armour}  & Exact \\
        $\pz{P}_1\pz{P}_2$ (PZ Multiplication) \cite[eq. (11)]{michaux2023armour}  & Exact \\
        $\setop{slice}(\pz{P}, \pzv_j, \sigma)$ \eqref{eq:pz_slice}  & Exact \\
        $\setop{sup}(\pz{P})$ \eqref{eq:pz_sup} and $\setop{inf}(\pz{P}) \eqref{eq:pz_inf}$  & Overapproximative \\
    \end{tabular}
    \label{tab:poly_zono_operations}
    \vspace*{-0.5cm}
\end{table}

There exists a variety of operations that one can perform on polynomial zonotopes such as minksowski sums, multiplying two or more polynomial zonotopes, and generating upper and lower bounds.
As illustrated in Tab. \ref{tab:poly_zono_operations}, the output of these operations is a polynomial zonotope that either exactly represents or over approximates the result of the operation on each individual element of the polynomial zonotope inputs.
For the interested reader, the operations given in Tab. \ref{tab:poly_zono_operations} are rigorously defined in \cite{michaux2023armour}.

\subsubsection{Interval Conversion}
Intervals can be represented as polynomial zonotopes.
Let $[z] = [\underline{z}, \overline{z}] \subset \mathbb{R}^n$, then $[z]$ can be converted to a polynomial zonotope $\pz{z}$ using
\begin{equation}
    \label{eq:int_to_pz}
    \pz{z} = \frac{\overline{z} + \underline{z}}{2} + \sum_{i = 1}^{n}\frac{\overline{z}_i - \underline{z}_i}{2}x_i,
\end{equation}
where $x \in [-1, 1]^n$ is the indeterminate vector.

\subsubsection{Slicing a PZ}  
One particularly useful property of polynomial zonotopes is the ability to obtain various subsets by plugging in different values of known indeterminates.
To this end, we introduce the ``slicing'' operation whereby a new subset is obtained from a polynomial zonotope $\pz{P} = \PZ{c, \pzgi, \pzei, \pzv, h_j, y_j}$ is obtained by evaluating one or more indeterminates.
Thus, given the $j$\ts{th} indeterminate $\pzv_j$ and a value $\sigma \in [-1, 1]$, slicing yields a subset of $\pz{P}$ by plugging $\sigma$ into the specified element $\pzv_j$:
\begin{equation}
    \label{eq:pz_slice}
   \hspace*{-0.25cm} \setop{slice}(\pz{P}, \pzv_j, \sigma) \subset \pz{P} =
        \left\{
            z \in \pz{P} \, \mid \, z = \sum_{i=0}^{\pzn} \pzgi \pzv ^{\pzei}, \, \pzv_j = \sigma
        \right\}.
\end{equation}

\subsubsection{Bounding a PZ}
In later sections, we also require operations to bound the elements of a polynomial zonotope.
We define the $\setop{sup}$ and $\setop{inf}$ operations, which return these upper and lower bounds, by setting the monomials $\pzv ^{\pzei} = 1$ and adding or subtracting all of the generators to the center, respectively.
For $\pz{P} \subseteq \R^n$, these operations return
\begin{align}
    \setop{sup}(\pz{P}) &= g_0 + \sum_{i=1}^{\pzn} \abs{\pzgi} + \sum_{j=1}^{n_h} \abs{h_j}, \label{eq:pz_sup}\\
    \setop{inf}(\pz{P}) &= g_0 - \sum_{i=1}^{\pzn} \abs{\pzgi} - \sum_{j=1}^{n_h} \abs{h_j}. \label{eq:pz_inf}
\end{align}
Note that these operations efficiently generate upper and lower bounds on the values of a polynomial zonotope through overapproximation.

\subsection{Distance Functions and Projections onto Sets}
The method we propose utilizes a signed distance function to compute the distance between the robot and obstacles:
\begin{defn}\label{subsec:def_sdf}
Given a subset $\Omega$ of $\mathbb{R}^{\nd}$, the \defemph{signed distance function} $\sdf$ between a point is a map $\sdf: \mathbb{R}^{\nd} \to \mathbb{R}$ defined as
    \begin{equation}
    \sdf(x; \Omega) = 
        \begin{cases}
          \dist(x; \partial\Omega)  & \text{if } x \in \Omega^{c} \\
          -\dist(x;\partial\Omega) & \text{if } x \in \Omega,
        \end{cases}
\end{equation}
where $\dist(x;\Omega)$ is the \defemph{distance function} associated with $\Omega$ and defined by:
    \begin{equation}
       \dist(x;\Omega) = \min_{y \in \partial\Omega} \|x - y\|.
    \end{equation}
\end{defn}
The exact signed distance function algorithm developed in Sec. \ref{subsec:sdf} utilizes Euclidean projections onto sets.
\begin{defn} For any $c \in \R^{\nd}$G, the \defemph{Euclidean projection} $\proj$ from $c$ to $\Omega \subset \R^{\nd}$ is defined as 
    \begin{equation}
    \proj(x; \Omega) = \{ \omega \in \Omega \;\: | \;\: \|x - \omega \| = \dist(x;\Omega)  \} .
\end{equation}
\end{defn}

\section{Arm and Environment Modeling}
\label{sec:modeling}
This section summarizes the environment, obstacles, and robot models used throughout the remainder of the paper. 

\subsection{Environment}\label{subsec:env}
\begin{figure}[t]
    \centering
    \includegraphics[width=0.95\columnwidth]{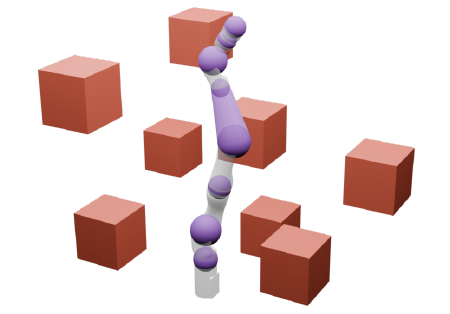}
    \caption{A visualization of the robot arm and its environment.
    The obstacles are shown in red and the robot is shown in grey (translucent).
    The volume of each joint is overapproximated by a sphere, shown in purple, in the workspace.
    Each link volume is overapproximated by a tapered capsule formed by the convex hull shown in light purple of two consecutive joint spheres (Assum. \ref{assum:joint_link_occupancy}).}
    \label{fig:assumption_figure}
    \vspace*{-0.3cm}
\end{figure}

The arm must avoid obstacles in the environment while performing safe motion planning. 
We begin by stating an assumption about the environment and its obstacles:
\begin{assum} \label{assum:env}
The robot and obstacles are all located in a fixed inertial world reference frame, which we denote $W \subset \R^{3}$.
We assume the origin of $\world$ is known and that each obstacle is represented relative to the origin of $\world$.
At any time, the number of obstacles $\nObs \in \N$ in the world is finite.
Let $\obsset = \{ O_1, O_2, \ldots, O_{\nObs} \}$ be the set of all obstacles.
Each obstacle is convex, bounded, and static with respect to time.
For each obstacle, we have access to a zonotope that overapproximates the obstacle's volume. See Fig. \ref{fig:assumption_figure}.
\end{assum}

\noindent Note that any convex, bounded object can always be overapproximated by a zonotope \cite{guibas2003zonotopes}.
In addition, any non-convex bounded obstacle can be outerapproximated by its convex hull.
For the remainder of document, safety is checked using the zonotope overapproximation of each obstacle.


\subsection{Arm Kinematics}\label{subsec:arm}
This work considers an $\nq$ degree of freedom serial robotic manipulator with configuration space $Q$.
Given a compact time interval $T \subset \R$, we define a trajectory for the configuration as $q: T \to Q \subset \R^{\nq}$ and a trajectory for the velocity as $\dot{q}: T \to \R^{\nq}$.
We make the following assumptions about the structure of the robot model:

\begin{assum}\label{assum:robot}
The robot operates in a three dimensional workspace, denoted $\workspace \subset \R^{3}$, such that $\workspace \subset \world$.
There exists a reference frame called the base frame, which we denote the $0$\ts{th} frame, that indicates the origin of the robot's kinematic chain.
The transformation between the robot's base frame and the origin of the world frame is known.
The robot is fully actuated and composed of only revolute joints, where the $j$\ts{th} joint actuates the robot's $j$\ts{th} link.
The robot's $j$\ts{th} joint has position and velocity limits given by $\qj \in [\qlim^-, \qlim^+]$ and $\qdj \in [\dqlim^-, \dqlim^+]$ for all $t \in T$, respectively. 
The robot's input is given by $u: T \to \R^{\nq}$.
\end{assum}

Recall from Assum. \ref{assum:robot} that the transformation from \textit{world} from to the the robot's $0$\ts{th}, or \textit{base}, frame is known.
We further assume that the $j$\ts{th} reference frame $\{\hat{x}_j, \hat{y}_j, \hat{z}_j\}$ is attached to the robot's $j$\ts{th} revolute joint, and that $\hat{z}_j = [0, 0, 1]^\top$ corresponds to the $j$\ts{th} joint's axis of rotation.

Then the forward kinematics $\FK_j: Q \to \R^{4 \times 4}$ maps a time-dependent configuration, $\q$, to the pose of the $j$\ts{th} joint in the world frame:
\begin{equation}\label{eq:fk_j}
    \FK_j(\q) = 
    \prod_{l=1}^{j} \homtrans_{l}^{l-1}(q_l(t)) = 
    \begin{bmatrix} R_j(\q) & p_j(\q) \\
    \zeros  & 1 \\
    \end{bmatrix},
\end{equation}
where 
\begin{equation}
    R_j(\q) \coloneqq R_j^{0}(\q) = \prod_{l=1}^{j}R_{l}^{l-1}(\ql)
\end{equation}
and 
\begin{equation}
    p_j(\q) = \sum_{l=1}^{j} R_{l}(\q) p_{l}^{l-1}.
\end{equation}

\noindent Note that we use the revolute joint portion of Assum. \ref{assum:robot} to simplify the description of forward kinematics; however, these assumptions and the methods described in the following sections can be readily be extended to other joint types such as prismatic or spherical joints in a straightforward fashion.
Note that the lack of input constraints means that one could apply an inverse dynamics controller \cite{Caccavale2020} to track any trajectory of of the robot perfectly.
It is also possible to deal with the tracking error due to uncertainty in the robot's dynamics \cite{michaux2023armour}.
However, this is outside of the scope of the current and as a result, we focus exclusively on modeling the kinematic behavior of the manipulator.

\subsection{Arm Occupancy}\label{subsec:arm_occupancy}
Next, we use the arm's kinematics to define the \textit{forward occupancy} which is the volume occupied by the arm in the workspace.
Let $L_j \subset \R^3$ denote the volume occupied by the $j$\ts{th} link with respect to the $j$\ts{th} reference frame. 
Then the forward occupancy of link $j$ is the map $\FO_j: Q \to \pow{\workspace}$ defined as
\begin{align}\label{eq:link_occupancy}
     \FO_j(\q) &= p_j(\q) \oplus R_j(\q) L_j,
\end{align}
where $p_j(\q)$ the position of joint $j$ and $R_j(\q) L_j$ is the rotated volume of link $j$.
The volume occupied by the entire arm in the workspace is given by the function $\FO: Q \to \workspace$ that is defined as
\begin{align}\label{eq:forward_occupancy}
    \FO(\q) = \bigcup_{j = 1}^{\nq} \FO_j(\q) \subset \workspace. 
\end{align}

Note that any of the robot's link volumes $L_j$ may be arbitrarily complex.
We therefore introduce an assumption to simplify the construction of the reachable set in Sec. \ref{subsec:sfo_construction}:
\begin{assum}\label{assum:joint_link_occupancy}
    For every $j \in \{1,\dots, \nq\}$ there exists a sphere of radius $r_j$ with center $p_j(\q)$ that overapproximates the volume occupied by the $j$\ts{th} joint in $\workspace$.
    Furthermore, the link volume $L_j$ is a subset of the tapered capsule formed by the convex hull of the spheres overapproximating the $j$\ts{th} and $j+1$\ts{th} joints.
    See Fig. \ref{fig:assumption_figure}.
\end{assum}
Following Assum. \ref{assum:joint_link_occupancy}, we define the sphere $\Sjq$ overapproximating the $j$\ts{th} joint as 
\begin{equation}\label{eq:joint_occupancy}
    \Sjq = B(p_j(\q), r_j)
\end{equation}
and the tapered capsule $TC_j(q(t))$ overapproximating the $j$\ts{th} link as
\begin{equation}
 TC_j(q(t)) = \conv\Big( \Sjq \cup \Sjpq \Big).
\end{equation}
Then, \eqref{eq:link_occupancy} and \eqref{eq:forward_occupancy} can then be overapproximated by
\begin{align}\label{eq:link_capsule_occupancy}
     \FO_j(\q)  &\subset \conv\Big( \Sjq \cup \Sjpq \Big) \nonumber \\
     &= TC_j(q(t)),
\end{align}
and
\begin{equation} \label{eq:forward_capsule_occupancy}
    \FO(\q) \subset \bigcup_{j = 1}^{\nq} TC_j(q(t)) \subset \workspace, 
\end{equation}
respectively.


For convenience, we use the notation $\FO(q(T))$ to denote the forward occupancy over an entire time interval $T$.
Note that we say that the arm is \textit{in collision} with an obstacle if $\FO_j(\q) \cap O_{\ell} \neq \emptyset$ for any $j \in \Nq$ or $\ell \in \NObs$ where $\NObs = \{1,\ldots,\nObs\}$.

\begin{algorithm}[t]
\small
\begin{algorithmic}[1]


\State $\pz{p_0}(\pzqi) \leftarrow \zeros$ 

\State$ \pz{R_0}(\pzqi) \leftarrow I_{3 \times 3}$ 

\State\hspace{0in}{\bf for} $j = 1:\nq$ 

    
    \State\hspace{0.2in}$\pz{R_{j}^{j-1}}(\pzqji), \pz{p_j}(\pzqi) \leftarrow \pz{\homtrans_{j}^{j-1}}(\pzqji)$ 
    
    \State\hspace{0.2in}$\pz{p_j}(\pzqi) \leftarrow \pz{p_{j-1}}(\pzqi) \oplus \pz{R_{j-1}}(\pzqi) \pz{p_{j}^{j-1}}$
        
    \State\hspace{0.2in}$ \pz{R_j}(\pzqi) \leftarrow  \pz{R_{j-1}}(\pzqi)  \pz{R_{j}^{j-1}}(\pzqji)$ 
    
    \State\hspace{0.2in}$\pz{\FK_j}(\pzqi) \leftarrow \{ \pz{R_j}(\pzqi) ,  \pz{p_j}(\pzqi) \}$ 
    
\State\hspace{0in}{\bf end for}


\end{algorithmic}
\caption{\small $\{\pz{\FK_j}(\pzqi)\,:\, j \in \Nq \} = \texttt{PZFK}(\pzqi)$}
\label{alg:compose_fk}
\end{algorithm}
\subsection{Trajectory Design}\label{subsec:trajectory_design}
\methodname{} computes safe trajectories  by solving an optimization program over parameterized trajectories at each planning iteration in a receding-horizon manner.
Offline, we pre-specify a continuum of trajectories over a compact set $K \subset \R^{n_k}$, $n_k \in \N$. 
Then each trajectory, defined over a compact time interval $T$, is uniquely determined by a \textit{trajectory parameter} $k \in K$.
Note that we design $K$ for safe manipulator motion planning, but $K$ may also be designed for other robot morphologies or to accommodate additional tasks \cite{holmes2020armtd, kousik2020bridging, kousik2019safe, liu2022refine} as long as it satisfies the following properties:
\begin{defn}[Trajectory Parameters]
\label{defn:traj_param}
For each $k \in K$, a \emph{parameterized trajectory} is an analytic function $q(\,\cdot\,; k) : T \to Q$ that satisfies the following properties:
\begin{outline}[enumerate]
\1 The parameterized trajectory starts at a specified initial condition $(q_0, \dot{q}_0)$, so that $q(0; k) = q_0$, and $\dot{q}(0; k) = \dot{q}_0$.
\1 Each parameterized trajectory brakes to a stop such that $\dot{q}(\tfin; k) = 0$.
\end{outline}
\end{defn}
\noindent \methodname{} performs real-time receding horizon planning by executing the desired trajectory computed at the previous planning iteration while constructing the next desired trajectory for the subsequent time interval.
Therefore, the first property ensures each parameterized trajectory is generated online and begins from the appropriate future initial condition of the robot.
The second property ensures that a braking maneuver is always available to bring the robot safely to a stop.

\begin{figure*}[ht]
    \centering
    \includegraphics[width=\textwidth]{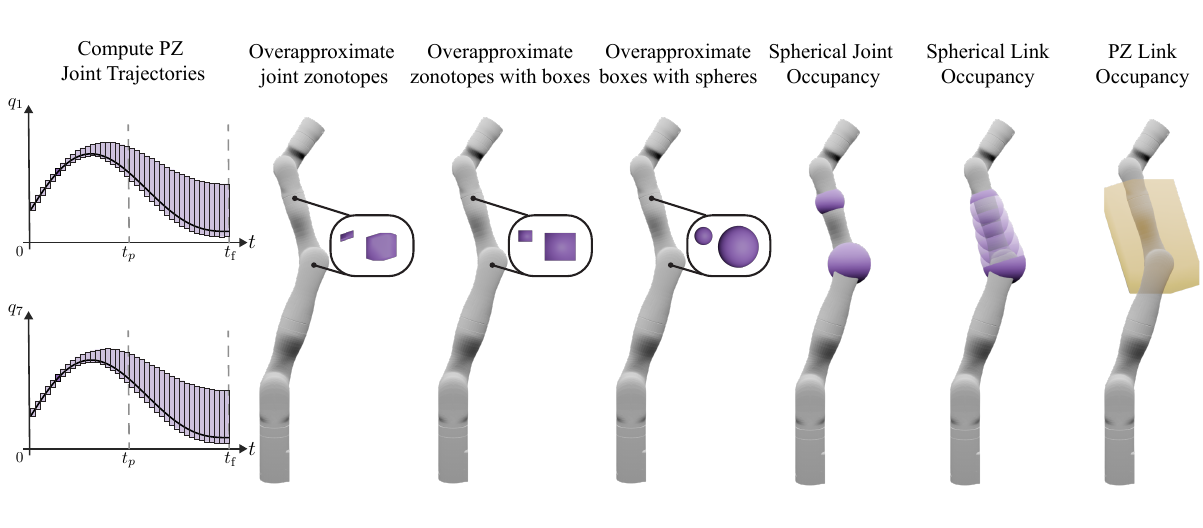}
    \caption{ 
    A visualization of the \SFO construction for a robotic arm in 3D.
    (First column) The planning time horizon is partitioned into a finite set of polynomial zonotopes (Sec. \ref{subsubsec:pz_time_horizon}) before computing the parameterized trajectory polynomial zonotopes shown in purple (Sec. \ref{subsubsec:pz_desired_traj}).
    The time horizon consists of a planning phase $[0,\tplan)$ and a braking phase $[\tplan, \tfin]$.
    A single desired trajectory for each joint is shown in black.
    (Second column) The polynomial zonotope forward kinematics algorithm (Sec. \ref{subsubsec:pzfk}) computes an overapproximation of the joint positions for each time interval.
    (Third column) The joint position zonotopes are first overapproximated by axis-aligned boxes and (Fourth column) subsequently overapproximated by the smallest sphere that circumscribes the box (Sec. \ref{subsubsec:sjo}).
    For the sake of clarity, the insets in columns 2 through 4 depict the polynomial zonotope joint positions, overapproximating boxes, and the circumscribed spheres as larger.
    (Fifth column) The radii of the circumscribed spheres are then added to the radii of the nominal joint spheres (Assum. \ref{assum:joint_link_occupancy}) to produce the \SJO (Lem. \ref{lem:sjo}).
    (Sixth column) Finally, the \SFO is computed for each link corresponding to the desired trajectory.
    (Seventh column) Note that \methodname generates link occupancies that are less conservative than \armtd.
    }\label{fig:sfo_3D}
\end{figure*}
\section{Planning Algorithm Formulation}\label{sec:planning}
The goal of this work is to construct collision free trajectories in a receding-horizon fashion by solving the following nonlinear optimization problem at each planning iteration:
\begin{align}
    \label{eq:optcost}
    &\underset{k\in K}{\min} &&\texttt{cost}(k) \\
    \label{eq:optpos}
    &&& q_j(t; k) \in [\qlim^-, \qlim^+]  &\forall t \in T, j \in \Nq \\
    \label{eq:optvel}
    &&& \dot{q}_j(t; k) \in [\dqlim^-, \dqlim^+]  &\forall t \in T,   j \in \Nq \\
    \label{eq:optcolcon}
    &&& \FO_j(q(t; k)) \bigcap \obsset = \emptyset  &\forall t \in T,   j \in \Nq
\end{align}
The cost function \eqref{eq:optcost} specifies a user- or task-defined objective, while each constraint guarantees the safety of any feasible parameterized trajectory.
The first two constraints ensure that the trajectory does not violate the robot's joint position \eqref{eq:optpos} and velocity limits \eqref{eq:optvel}.
The last constraint \eqref{eq:optcolcon} ensures that the robot does not collide with any obstacles in the environment.

Implementing a real-time algorithm to solve this optimization problem is challenging for two reasons.
First, obstacle-avoidance constraints are typically non-convex \cite{borrelli2020collision}. 
Second, each constraint must be satisfied for all time $t$ in a continuous, and uncountable, time interval $T$.
To address these challenges, this section develops \methodname a novel, real-time trajectory planning method that combines reachability analysis with differentiable collision primitives. 
We first discuss how to overapproximate a family of parameterized trajectories that the robot follows.
Next, we discuss how to overapproximate \eqref{eq:forward_capsule_occupancy} using the Spherical Forward Occupancy ($\sfo$).
Then, we discuss how to compute the signed distances between the forward occupancy and obstacles in the robot's environment.
Finally, in Sec. \ref{subsec:online_planning} we discuss how to combine the $\sfo$ and signed distance function to generate obstacle-avoidance constraints for a real-time algorithm that can be used for safe online motion planning.

\subsection{Polynomial Zonotope Trajectories and Forward Kinematics}\label{subsec:pz_traj}
This subsection summarizes the relevant results from \cite{michaux2023armour} that are used to overapproximate parameterized trajectories (Def. \ref{defn:traj_param}) using polynomial zonotopes.

\subsubsection{Time Horizon and Trajectory Parameter PZs}\label{subsubsec:pz_time_horizon}
We first describe how to create polynomial zonotopes representing the planning time horizon $T$.
We choose a timestep $\timestep$ so that $\nt \coloneqq \frac{T}{\timestep} \in \N$ divides the compact time horizon $T \subset \R$ into $\nt$ time subintervals denoted by $N_t := \{1,\ldots,\nt\}$.
We then represent the $i$\ts{th} time subinterval, corresponding to $t \in \iv{(i-1)\timestep, i\timestep}$, as a polynomial zonotope $\pz{T_i}$, where 
\begin{equation}
    \label{eq:time_pz}
    \pz{T_i} =
        \left\{t \in T \mid 
            t = \tfrac{(i-1) + i}{2}\timestep + \tfrac{1}{2} \timestep \tvari,\ \tvari \in [-1,1]
        \right\}
\end{equation}
with indeterminate $\tvari \in \iv{-1, 1}$.
We also use polynomial zonotopes to represent the set of trajectory parameters $K$.
For each $j \in \{1, \dots, \nq\}$, $\Kj$ is the compact one-dimensional interval  $\Kj = \iv{-1, 1}$ such that $K$ is given by the Cartesian product $K = \bigtimes_{i=1}^{n_q} K_i$.
Following \eqref{eq:int_to_pz}, we represent the interval $\Kj$ as a polynomial zonotope $\pz{\Kj} = \kjvar$ where $\kjvar \in \iv{-1, 1}$ is an indeterminate.

\subsubsection{Parameterized Trajectories}\label{subsubsec:pz_desired_traj}
Using the time partition $\pz{T_i}$ and trajectory parameter $\pz{K_j}$ polynomial zonotopes described above, we create polynomial zonotopes $\pzqji$ and $\pzqdji$ that overapproximate the parameterized position (Fig. \ref{fig:sfo_3D}, col. 1) and velocity trajectories defined in Def. \ref{defn:traj_param} for all $t$ in the $i$\ts{th} time subinterval. 
This is accomplished by plugging $\pz{T_i}$ and $\pz{K}$ into the formulas for $\qkj$ and $\qdkj$.
For convenience, we restate \cite[Lemma 14]{michaux2023armour} which proves that sliced position and velocity polynomial zonotopes are overapproximative:
\begin{lem}[Parmaeterized Trajectory PZs]
\label{lem:pz_desired_trajectory}
The parameterized trajectory polynomial zonotopes $\pzqji$ are overapproximative, i.e., for each $j \in \Nq$ and $k\in \pz{K_j},$ 
\begin{equation}
    \qkj \in \pzqjki \quad \forall t \in \pz{T_i}
\end{equation}
One can similarly define $\pzqdji$ that are also overapproximative. 
\end{lem}

\subsubsection{PZ Forward Kinematics}\label{subsubsec:pzfk}
We begin by representing the robot's forward kinematics \eqref{eq:fk_j} using polynomial zonotope overapproximations of the joint position trajectories.
Using $\pzqi$, we compute $\pz{p_j}(\pzqi)$ and $\pz{R_j}(\pzqi)$, which represent overapproximations of the position and orientation of the $j$\ts{th} frame with respect to frame $(j-1)$ at the $i$\ts{th} time step.
The polynomial zonotope forward kinematics can be computed (Alg. \ref{alg:compose_fk}) in a similar fashion to the classical formulation \eqref{eq:fk_j}.
For convenience, we restate \cite[Lemma 17]{michaux2023armour} that proves that this is overapproximative:
\begin{lem}[PZ Forward Kinematics]
\label{lem:PZFK}
Let the polynomial zonotope forward kinematics for the $j$\ts{th} frame at the $i$\ts{th} time step be defined as
\begin{equation}\label{eq:pzfk_j}
    \pzFKjKi = 
    \begin{bmatrix} \pz{R_j}(\pzqi) & \pz{p_j}(\pzqi)\\
    \mathbf{0}  & 1 \\
    \end{bmatrix},
\end{equation}
where
\begin{align}
    \pz{R_j}(\pzqi) &= \prod_{l=1}^{j} \pz{R_{l}^{l-1}}(\pzqli), \\
    \pz{p_j}(\pzqi) &= \sum_{l=1}^{j} \pz{R_l}(\pzqi) \pz{p_{l}^{l-1}},
\end{align}
then for each  $j \in \Nq$,  $k \in \pz{K}$,  $\FK_j(q(t;k)) \in  \pzFKjki$ for all $t \in \pz{T_i}$.
\end{lem}

\subsection{\SFO Construction}\label{subsec:sfo_construction}
This subsection describes how to use the parameterized trajectory polynomial zonotopes discussed in Section \ref{subsec:pz_traj} to construct the \SFO{} introduced in Section \ref{subsubsec:sfo}.
\subsubsection{\SJO}\label{subsubsec:sjo}
We next introduce the \SJO to compute an overapproximation of the spherical volume occupied by each arm joint in the workspace.
In particular, we show how to construct a sphere that overapproximates the volume occupied by the $j$\ts{th} joint over the $i$\ts{th} time interval $\pz{T_i}$ given a parameterized trajectory polynomial zonotope $\pzqi$.
First we use the polynomial zonotope forward kinematics to compute $\pz{p_j}(\pzqi)$, which overapproximates the position of the $j$\ts{th} joint frame (Fig. \ref{fig:sfo_3D} col. 2).
Next, we decompose $\pz{p_j}(\pzqi)$ into
\begin{equation}
    \pz{p_j}(\pzqi) = \pzCjiK \oplus \pzuj,
\end{equation}
where $\pzCjiK$ is a polynomial zonotope that only contains generators that are dependent on $\pz{K}$ and $\pzuj$ is a zonotope that is independent of $\pz{K}$.
We then bound $\pzuj$ with an axis aligned cube (Fig. \ref{fig:sfo_3D}, col. 3) that is subsequently bounded by a sphere (Fig. \ref{fig:sfo_3D}, col. 4) of radius $u_{j,i}$ .
Finally, the \SJO is defined as
\begin{equation}
    \SjiK = \pzCjiK \oplus B(\mathbf{0}, r_j + u_{j,i}),
\end{equation}
where $\pzCjiK$ is a polynomial zonotope representation of the sphere center, $r_j$ is the radius nominal joint sphere, and $u_{j,i}$ is the sphere radius that represents additional uncertainty in the position of the joint.
Because $\pzCjiK$ is only dependent on $\pz{K}$, slicing a particular trajectory parameter $k \in K$ yields a single point for the sphere center $\pzCjik$. 
We state this result in the following lemma whose proof is in Appendix \ref{appendix:sjo}:

\begin{lem}[Spherical Joint Occupancy]\label{lem:sjo}
Let the \emph{spherical joint occupancy} for the 
$j$\ts{th}  frame at the $i$\ts{th} time step be defined as
\begin{equation}
    \SjiK =\pzCjiK \oplus B(\mathbf{0}, r_j + u_{j,i}),
\end{equation}
where $\pzCjiK$ is the center of $\pz{p_j}(\pzqi)$, $r_{j}$ is the radius of $\Sjqk$, and $u_{j,i}$ is the radius of a sphere that bounds the independent generators of $\pz{p_j}(\pzqi)$.
Then,
\begin{equation}
    \Sjqk) \subset \Sjik
\end{equation}
for each  $j \in \Nq$,  $k \in \pz{K}$,  $\Sjqk \in  \Sjik$ for all $t \in \pz{T_i}$.
\end{lem}

\subsubsection{\SFO}\label{subsubsec:sfo}

Previous work \cite{michaux2023armour} used \eqref{eq:link_occupancy} along with Lem. \ref{lem:PZFK} to overapproximate the forward occupancy of the $j$\ts{th} link over the $i$\ts{th} time step using polynomial zonotopes:
\begin{align}\label{eq:pz_forward_occupancy_j}
     \pzFOjKi = \pz{p_j}(\pzqi) \oplus \pz{R_j}(\pzqi)\pz{L_j}.
\end{align}

Our key insight is replacing \eqref{eq:link_occupancy} with an alternate representation based on a tapered capsules \eqref{eq:link_capsule_occupancy}.
Recall from Assum. \ref{assum:joint_link_occupancy} that the $j$\ts{th} link can be outer-approximated by the tapered capsule formed by the two spheres encompassing the closest joints.
Sec. \ref{subsubsec:sjo} described how to overapproximate each joint volume over a continuous time interval $\pz{T_i}$.
Therefore, by Assum. \ref{assum:joint_link_occupancy}, the volume occupied by $j$\ts{th} link over the $\pz{T_i}$ is overapproximated by the tapered capsule formed by $\Sjik$ and $\Sjpik$ for a given trajectory parameter $k \in K$:
\begin{equation}
    \TCjik = \conv(\Sjik \cup \Sjpik) \quad \forall t \in \pz{T_i}.
\end{equation}

However, tapered capsules are not ideal for performing collision checking with arbitrary, zonotope-based obstacles.
Instead, we introduce the \SFO ($\sfo$) to overapproximate $\TCjik$ with $\ns \in \N$ spheres (Fig. \ref{fig:sfo_3D}, col. 6):

\begin{thm}[Spherical Forward Occupancy]\label{thm:sfo}
Let $\ns \in \N$ be given and let the \SFO for the $j$\ts{th} link at the $i$\ts{th} time step be defined as
\begin{equation}\label{eq:tapered_capsule_overapprox}
    \sfojk = \left\{\Sbarjimk \, : \, m \in N_s \right\},
\end{equation}
where $N_s = \{1, \cdots, \ns\}$ and
\begin{equation}
    \Sbarjimk = B\left(\bar{c}_{j,i,m}(k), \bar{r}_{j,i,m}(k)\right),
\end{equation}
is the $m$\ts{th} link sphere with with center $\bar{c}_{j,i,m}(k)$ and radius $\bar{r}_{j,i,m}(k)$.
Then
\begin{equation}
    \TCjik \subset \sfojk
\end{equation}
for each  $j \in \Nq$,  $k \in \pz{K}$, and $t \in \pz{T_i}$.
\end{thm}
Theorem \ref{thm:sfo} yields a tighter reachable set (Fig. \ref{fig:sfo_3D}, col. 6) compared to previous methods (Fig. \ref{fig:sfo_3D}, col. 7) while remaining overapproximative. 
Sec. \ref{subsec:online_planning} describes how the \SFO is used to compute collision-avoidance constraints.

\begin{algorithm}[t]
\begin{algorithmic}[1]
\State $A, b, E \leftarrow \mathcal{O}$
\State $d_{\text{face}} \leftarrow \max(Ac-b)$  
\State{\bf if} $d_{\text{face}} \leq 0$ // check negative distances to faces
\State\hspace{0.2in} \textbf{Return} $d_{\text{face}}$ 

\State{\bf else} // check positive distances to faces
\State\hspace{0.2in}{\bf for} $i_A = 1:m_A$ 
    \State\hspace{0.4in} $d_{\text{face}} \leftarrow (A c- b)_{i_A}$
    \State\hspace{0.4in} $p_{\text{face}} \leftarrow c - d_{\text{face}} \cdot A_{i_A}^{T} $ 
    \State\hspace{0.4in}{\bf if} $d_{\text{face}} \geq 0$ \textbf{and} $ \max(A p_{\text{face}} - b) \leq 0$ 
        \State\hspace{0.6in} \textbf{Return} $d_{\text{face}}$
    \State\hspace{0.4in}{\bf end if}
    \State\hspace{0.2in}{\bf end for}
\State{\bf end if}
\State $d_{\text{edge}} \leftarrow \inf$
\State{\bf for} $i_E = 1: |E|$ // check distances to edges
\State\hspace{0.2in} $d_{\text{edge}} \leftarrow \min(\dist(c,E_{i_E}), d_{\text{edge}})$
\State{\bf end for}
\State \textbf{Return} $d_{\text{edge}}$
\end{algorithmic}
\caption{\SDF$(\mathcal{O}, c)$ }
\label{alg:sdf}
\end{algorithm}

\subsection{Exact Signed Distance Computation}\label{subsec:sdf}
The Spherical Forward Occupancy constructed in \ref{subsubsec:sfo} is composed entirely of spheres.
A sphere can be thought of as a point with a margin of safety corresponding to its radius.
Therefore, utilizing spheres within obstacle-avoidance constraints requires only point-to-obstacle distance calculations.
To make use of this sphere-based representation for trajectory planning, this subsection presents Alg. \ref{alg:sdf} for computing the exact signed distance between a point and a zonotope in 3D based on the following lemma whose proof can be found in the appendix:
\begin{lem}\label{lem:sdf}
Given an obstacle zonotope $\mathcal{O} \in \R^3$, consider its polytope representation with normalized rows $(A,b)$ and edges $E$ computed from its vertex representation.
Then for any point $c \in \R^3$,
\begin{equation}
    \sdf(c; \mathcal{O}) = 
        \begin{cases}
          \max(Ac-b)  & \text{if } c \in \mathcal{O} \\
          (Ac-b)_{i_A} & \text{if } c \not \in \mathcal{O}, \max(A p_{\text{face}} - b) \leq 0 \\
          \underset{E_{i_E} \in E}{\min} \dist(c; E_{i_E})  & \text{otherwise}
        \end{cases}    ,
\end{equation}
$i_A$ indexes the rows of $(Ac-b)$, $i_E$ indexes the edges of $E$, and $p_{\text{face}} = c - (Ac-b)_{i_A} \cdot A_{i_A}^{T}$ is the projection of $c$ onto the $i_A$\ts{th} hyperplane of $\mathcal{O}$ whose normal vector is given by $A_{i_A}^{T}$.
\end{lem}

\subsection{Trajectory Optimization}\label{subsec:trajectory_optimization}
We combine parameterized polynomial zonotope trajectories (Sec. \ref{subsec:pz_traj}) with the \SFO (Sec. \ref{subsubsec:sfo}) and an exact signed distance function (Sec. \ref{subsec:sdf}) to formulate the following trajectory optimization problem:
\begin{align}
    \label{eq:pzoptcost}
    &\underset{k \in K}{\min} &&\numop{cost}(k) \hspace{3cm} \opt&  \\
    \label{eq:pz_optpos}
    &&& \hspace*{-0.75cm} \pzqjki \subseteq [\qlim^-, \qlim^+]  \hspace{1cm} \forall (i,j) \in N_t \times N_q \\
    \label{eq:pz_optvel}
    &&& \hspace*{-0.75cm}\pzqdjki \subseteq [\dqlim^-, \dqlim^+]  \hspace{1cm} \forall (i,j) \in N_t \times N_q \\
    \label{eq:pz_optcolcon}
    &&& \hspace*{-0.75cm}\texttt{SDF}(\mathcal{O}_{n}, \bar{c}_{j,i,m}(k)) > \bar{r}_{j,i,m}(k) \nonumber\\
    &&& \hspace{2.2cm} \forall (i,j,m,n) \in N_t \times N_q \times N_s \times N_{\mathcal{O}},
\end{align}
where $i \in N_t$ indexes the time intervals $\pz{T_i}$, $j \in N_q$ indexes the $j$\ts{th} robot joint, $m \in N_s$ indexes the spheres in each link forward occupancy, and $n \in N_{\mathcal{O}}$ indexes the number of obstacles in the environment.

\subsection{Safe Motion Planning}\label{subsec:online_planning}
\methodname's planning algorithm is summarized in Alg. \ref{alg:online_planning}.
The polytope representation $(A,b)$ and edges $E$ for computing the signed distance (Alg. \ref{alg:sdf}) to each obstacle $\mathcal{O}_n \in \obsset$ are computed offline before planning.
Every planning iteration is given $\tplan$ seconds to find a safe trajectory.
The $\sjo$ (Lem. \ref{lem:sjo}) is computed at the start of every planning iteration outside of the optimization problem and the $\sfo$ (Thm. \ref{thm:sfo}) is computed repeatedly while numerically solving the optimization problem. If a solution to \optref{} is not found, the output of Alg. \ref{alg:online_planning} is set to \nan and the robot executes a braking maneuver (Line \ref{lin:brake}) using the trajectory parameter found in the previous planning iteration.
To facilitate real-time motion planning, analytical constraint gradients for \eqref{eq:pz_optpos}--\eqref{eq:pz_optcolcon} are provided to aid computing the numerical solution of \optref.

\begin{algorithm}[t]
\small
\begin{algorithmic}[1]
    \State {\bf Require:} $\tplan > 0$, $\Nt, N_s \in \N, \obsset, \SDF$, and $\texttt{cost}: K \to \R$.
    \State {\bf Initialize:} $i = 0$, $t_i = 0$, and \newline
    \phantom{\bf stupid} $\{ k^*_j \} = \texttt{Opt}(\qstart,\zeros,\zeros,\obsset,\texttt{cost}, \tplan, \Nt, N_s, \SDF, \sfo)$
    \State {\bf If} $k^*_{j} = \texttt{NaN}$, {\bf then} break
    \State{\bf Loop:}
    
        \State \hspace{0.05in} // Line \ref{lin:apply} executes simultaneously with Lines \ref{lin:opt} -- \ref{lin:else} //
        \State \hspace{0.05in}  // Use Lem. \ref{lem:sjo}, Thm. \ref{thm:sfo}, and Alg. \ref{alg:sdf}//  
        
        \hspace{0.05in} {\bf Execute} $q_d(t;k^*_i)$  on robot for $t \in [t_i,t_i + \tplan]$ \label{lin:apply} 
      \State \hspace{0.05in}  $\{ k^*_{i+1} \} = \texttt{Opt}(q_d(\tplan;k^*_j), \dot{q}^{j}_d(\tplan;k^*_j), \ddot{q}^{j}_d(\tplan;k^*_j),\obsset,$ \phantom{stupid stupid stupid}$\texttt{cost},\tplan, \Nt, N_s, \SDF, \sfo)$ \label{lin:opt} 
        \State \hspace{0.05in} {\bf If} $k^*_{i+1} = \texttt{NaN}$, {\bf then} break
        \State \hspace{0.05in} {\bf Else}  $t_{i+1} \leftarrow t_i + \tplan$ and $i \leftarrow i + 1$ \label{lin:else}

    \State{\bf End}
    \State {\bf Execute} $q_d(t;k^*_i)$  on robot for $t \in [t_i+\tplan,t_i + \tfin]$ \label{lin:brake}
\end{algorithmic}
\caption{\small \methodname{} Online Planning}
\label{alg:online_planning}
\end{algorithm}

\section{Demonstrations}\label{sec:demonstrations}
We demonstrate \methodname in simulation using the Kinova Gen3 7 DOF manipulator.

\subsection{Implementation Details}
Experiments for \methodname, \armtd \cite{holmes2020armtd}, and \mpot \cite{le2023mpot} were conducted on a computer with an Intel Core i7-8700K CPU @ 3.70GHz CPU and two NVIDIA RTX A6000 GPUs. 
Baseline experiments for \chomp \cite{Zucker2013chomp} and \trajopt \cite{Schulman2014trajopt} were conducted on a computer with an Intel Core i9-12900H @ 4.90GHz CPU.
Polynomial zonotopes and their arithmetic are implemented in PyTorch \cite{paszke2017automatic} based on the CORA toolbox \cite{cora2018}.

\subsubsection{Simulation and Simulation Environments}
We use a URDF of the Kinova Gen3 7-DOF serial manipulator \cite{kinova-user-guide} from \cite{kinovakortex} and place its base at the origin of each simulation environment.
Collision geometry for the robot is provided as a mesh, which we utilize with the trimesh library \cite{trimesh} to check for collisions with obstacles. 
Note that we do not check self-collision of the robot.
For simplicity, all obstacles are static, axis-aligned cubes parameterized by their centers and edge lengths. 
\subsubsection{Signed Distance Computation}
Recall from Assum. \ref{assum:env} that each obstacle is represented by a zonotope $Z$.
The edges $E$ and polytope representation $(A,b)$ of $Z$ are computed offline before planning begins.
PyTorch is used to build the computation graph for calculating the signed distances in Alg. \ref{alg:sdf}.
The analytical gradients of the output of Alg. \ref{alg:sdf} are derived and implemented to speed up the optimization problem \optref.

\subsubsection{Desired Trajectory}
We parameterize the trajectories with a piece-wise linear velocity such that the
 \defemph{trajectory parameter} $k=(k_1,\cdots,k_{\nq})\in\R^{\nq}$ is a constant acceleration over the planning horizon $[0, \tplan)$.
The trajectories are also designed to contain a constant braking acceleration applied over $[\tplan, \tfin]$ that brings the robot safely to a stop.
Thus, given an initial position $q_{0}$ and velocity $\dot{q}_{0}$, the parameterized trajectory is given by:
\begin{align}\label{eq:actual_trajectory}
    q(t;k) = \begin{cases}
        q_0 + \dot{q}_0t + \frac{1}{2}k\,t^2, &t \in [0,\tplan) \\
        q_0 + \dot{q}_0\tplan + \frac{1}{2}k\,\tplan^2 + \\ (\dot{q}_0\tplan + k\tplan)\frac{(2\tfin-\tplan-t)(t-\tplan)}{2(\tfin - \tplan)}, &t \in [\tplan,\tfin].
    \end{cases}
\end{align}

\subsubsection{Optimization Problem Implementation}
\methodname{} and \armtd{} utilize IPOPT \cite{ipopt-cite} for online planning by solving \optref.
The constraints for \optref, which are constructed from the \SJO (Lem. \ref{lem:sjo}) and \SFO (Thm. \ref{thm:sfo}), are implemented in PyTorch using polynomial zonotope arithmetic based on the CORA toolbox \cite{cora2018}.
All of the constraint gradients of \optref{} were computed analytically to improve computation speed.
Because polynomial zonotopes can be viewed as polynomials, all of the constraint gradients can be readily computed using the standard rules of differential calculus.

\subsubsection{Comparisons}
We compare \methodname to \armtd \cite{holmes2020armtd} where the reachable sets are computed with polynomial zonotopes similar to \cite{michaux2023armour}.
We also compare to \chomp \cite{Zucker2013chomp}, \trajopt \cite{Schulman2014trajopt}, and \mpot \cite{le2023mpot}.
We used the ROS MoveIt \cite{moveit2014} implementation for \chomp, Tesseract Robotics's ROS package \cite{tesseracttrajopt} for \trajopt, and the original implementation for \mpot \cite{le2023mpot}.
Note that \mpot's signed distance function was modified to account for the obstacles in the planning tasks we present here.
Although \chomp, \trajopt, and \mpot are not receding-horizon planners, each method provides a useful baseline for measuring the difficulty of the scenarios encountered by \methodname.

\begin{figure*}[ht]
    \centering
    \includegraphics[width=\textwidth]{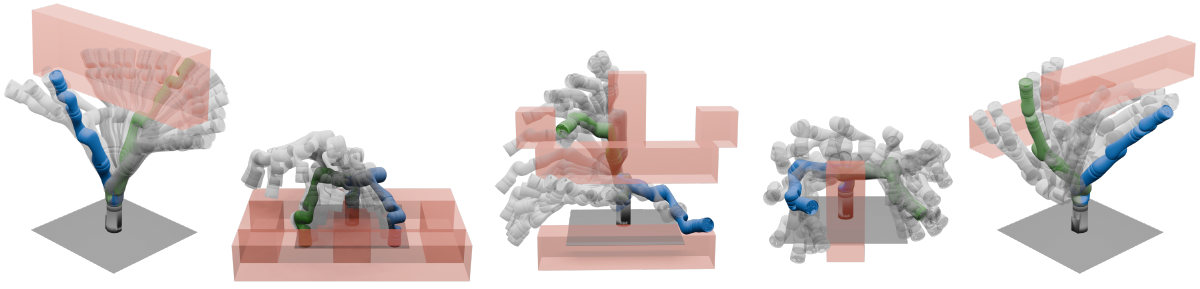}
    \caption{Subset of \textbf{Hard Scenarios} where \methodname succeeds.
    The start, goal, and intermediate poses are shown in blue, green, and grey (transparent), respectively. 
    Obstacles are shown in red (transparent).
    The tasks include reaching from one side of a wall to another, between two small bins, from below to above a shelf, around one vertical post, and between two horizontal posts. }
    \label{fig:hard_scenarios}
\end{figure*}
\subsection{Runtime Comparison with \armtd}

\begin{table}[t]
        \centering
            \begin{tabular}{ | c || c | c | c ||}
                \hline 
                Methods & \multicolumn{3}{c||}{mean constraint evaluation time [ms]} \\\hline
                \# Obstacles (s) & 10 & 20 & 40 \\\hline \hline
                \methodname{} $(\pi/24)$ & \textbf{3.1 ± 0.2} & \textbf{3.8 ± 0.1} & \textbf{5.1 ± 0.1}\\ \hline
                ARMTD $(\pi/24)$          & 4.1 ± 0.3 & 5.4 ± 0.5 & 8.3 ± 0.7 \\ \hline
                \methodname{} $(\pi/6)$   & \textbf{3.1 ± 0.1} & \textbf{3.8 ± 0.1} & 5.2 ± 0.1 \\ \hline
                ARMTD $(\pi/6)$           & 4.1 ± 0.3 & 5.4 ± 0.5 & 8.1 ± 0.7 \\ \hline
                \end{tabular}
        \caption{Mean runtime for constraint and constraint gradient evaluation for \methodname{} and \armtd in Kinova planning experiments with 10, 20, and 40 obstacles under 0.5s time limit $\downarrow$}
        \label{tab:consevalruntime3d7link0.5s}
\end{table}

\begin{table}[t]
        \centering
            \begin{tabular}{ | c || c | c | c ||}
                \hline 
                Methods & \multicolumn{3}{c||}{mean planning time [s]} \\\hline
                \# Obstacles (s) & 10 & 20 & 40 \\\hline \hline
                \methodname{} $(\pi/24)$ & \textbf{0.14 ± 0.07} & \textbf{0.16 ± 0.08} & \textbf{0.20 ± 0.09}\\ \hline
                ARMTD $(\pi/24)$         & 0.18 ± 0.08 & 0.30 ± 0.10 & 0.45 ± 0.08 \\ \hline
                \methodname{} $(\pi/6)$   & 0.16 ± 0.08 & 0.18 ± 0.08 & 0.24 ± 0.09 \\ \hline
                ARMTD $(\pi/6)$           & 0.24 ± 0.09 & 0.38 ± 0.10 & \textcolor{red}{0.51 ± 0.04} \\ \hline
                \end{tabular}
        \caption{Mean per-step planning time for \methodname{} and ARMTD in Kinova planning experiments with 10, 20, and 40 obstacles under 0.5s time limit $\downarrow$. \textcolor{red}{Red} indicates that the average planning time limit has been exceeded.}
        \label{tab:planningtime3d7link0.5s}
\end{table}

\begin{figure}[t]
    \centering
    \includegraphics[width=0.93\columnwidth]{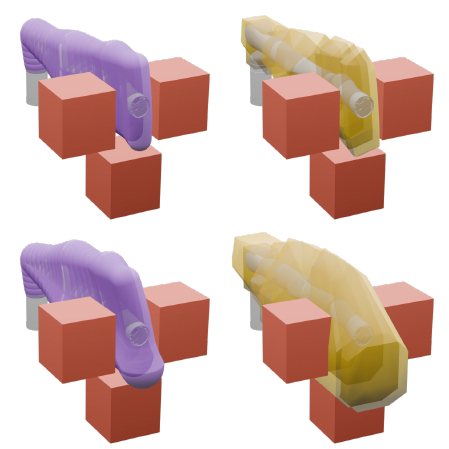}
    \caption{A comparison of \methodname' and \armtd's reachable sets. 
    (Top row) Both \methodname and \armtd can generate trajectories that allow the arm to fit through a narrow passage when the maximum acceleration is limited to $\frac{\pi}{24}$ rad/s$^2$. 
    (Bottom row) However, when the maximum acceleration is $\frac{\pi}{6}$ rad/s$^2$ \armtd's reachable set collides with the obstacles while \methodname' remains collision-free.}
    \label{fig:reachset_comparison}
    \vspace*{-0.3cm}
\end{figure}

We compare the runtime performance of \methodname{} to \armtd{} while varying the number of obstacles, maximum acceleration, and planning time limit.
We measure the mean constraint evaluation time and the mean planning time.
Constraint evaluation includes the time to compute the constraints and the constraint gradients.
Planning time includes the time required to construct the reachable sets and the total time required to solve \optref{} including constraint gradients evaluations.
We consider planning scenarios with $n=10$, $20$, and $40$ obstacles each placed randomly such that each obstacle is reachable by the end effector of the robot.
We also consider two families of parameterized trajectories \eqref{eq:actual_trajectory} such that the maximum acceleration is $\pi/24$ rad/s$^2$ or $\pi/6$ rad/s$^2$, respectively.
The robot is allowed to have $7$, $14$, or $21$ degrees of freedom by having 1, 2, or 3 Kinova arms in the environment.
The  maximum planning time is limited to $0.15$s, $0.25$s or $0.5$s for the $7$ DOF arm and $0.5$s and $1.0$s for the $14$ and $21$ DOF arms.
For each case, the results are averaged over 100 trials.

Tables \ref{tab:consevalruntime3d7link0.5s} and \ref{tab:planningtime3d7link0.5s} summarize the results for a $7$ DOF robot with a $0.5$s planning time limit.
\methodname$(\pi/24)$ has the lowest mean constraint evaluation time and per-step planning time followed by \methodname$(\pi/6)$. 
For each maximum acceleration condition, \methodname' average planning time is almost twice as fast as \armtd when there are 20 or more obstacles.
Note that we see a similar trend when the planning time is limited to $0.25s$ and $0.15s$.
Further note that under planning time limits of $0.25$s (Table \ref{tab:consevalruntime3d7link0.25s}) and $0.15$s (Table \ref{tab:consevalruntime3d7link0.15s}), \armtd is unable to evaluate the constraints quickly enough to solve the optimization problem within the alloted time as the number of obstacles increases \ref{tab:consevalruntime3d7link0.25s}. 
This is consistent with the cases when \armtd's mean planning time limit exceeds the maximum allowable planning time limits in Tables \ref{tab:planningtime3d7link0.5s}, \ref{tab:planningtime3d7link0.25s}, and \ref{tab:planningtime3d7link0.15s}.

\begin{table*}[t]
        \centering
            \begin{tabular}{ | c || c | c | c || c | c  | c ||}
                \hline 
                Methods & \multicolumn{6}{c||}{mean constraint evaluation time [ms]}\\\hline
                \# DOF & \multicolumn{3}{c||}{14} & \multicolumn{3}{c||}{21} \\  \hline
                \# Obstacles & 5 & 10 & 15 & 5 & 10 & 15 \\ \hline \hline
                \methodname{} $(\pi/24)$ & \textbf{3.9 ± 0.2} & \textbf{4.6 ± 0.2} & \textbf{5.4 ± 0.1} & \textbf{4.8 ± 0.2} & \textbf{6.0 ± 0.1} & \textbf{7.2 ± 0.1} \\ \hline
                ARMTD $(\pi/24)$ & 8.4 ± 0.6 & 9.4 ± 0.7 & 10.7 ± 0.8 & 12.9 ± 0.9 & 14.5 ± 0.7 & 16.1 ± 0.8  \\ \hline
                \methodname{} $(\pi/6)$ & \textbf{3.9 ± 0.3} & \textbf{4.6 ± 0.2} & \textbf{5.4 ± 0.1} & \textbf{4.8 ± 0.1} & \textbf{6.0 ± 0.1} & \textbf{7.2 ± 0.1} \\ \hline
                ARMTD $(\pi/6)$ & 8.4 ± 0.6 & 9.4 ± 0.7 & 10.6 ± 0.7 & 12.8 ± 0.8 & 14.5 ± 0.7 & 16.7 ± 1.4  \\ \hline
                \end{tabular}
        \caption{Mean runtime for constraint and constraint gradient evaluation for \methodname{} and ARMTD in 2 and 3 Kinvoa arms planning experiment with 5, 10, 15 obstacles under 1.0s time limit $\downarrow$}
        \label{tab:multiarm_consevaltime_1.0s}
\end{table*}

\begin{table*}[t]
        \centering
            \begin{tabular}{ | c || c | c | c || c | c | c | c ||}
                \hline 
                Methods & \multicolumn{6}{c||}{mean planning time [ms]}\\\hline
                \# DOF & \multicolumn{3}{c||}{14} & \multicolumn{3}{c||}{21} \\  \hline
                \# Obstacles & 5 & 10 & 15 & 5 & 10 & 15 \\ \hline \hline
                \methodname{} $(\pi/24)$ & \textbf{0.33 ± 0.10} & \textbf{0.37 ± 0.14} & \textbf{0.41 ± 0.16} & \textbf{0.63 ± 0.14} & \textbf{0.67 ± 0.16} & \textbf{0.71 ± 0.16}  \\ \hline
                ARMTD $(\pi/24)$ & 0.38 ± 0.10 & 0.57 ± 0.16 & 0.76 ± 0.17 & 0.70 ± 0.13 & 0.98 ± 0.07 & \textcolor{red}{1.07 ± 0.03}\\ \hline
                \methodname{} $(\pi/6)$ & 0.38 ± 0.12 & 0.46 ± 0.16 & 0.50 ± 0.17 & 0.72 ± 0.16 & 0.79 ± 0.17 & 0.85 ± 0.16 \\ \hline
                ARMTD $(\pi/6)$ & 0.49 ± 0.16 & 0.72 ± 0.18 & 0.91 ± 0.15 & 0.81 ± 0.15 & \textcolor{red}{1.03 ± 0.03} & \textcolor{red}{1.06 ± 0.03}   \\ \hline
                \end{tabular}
        \caption{Mean per-step planning time for \methodname{} and ARMTD in 2 and 3 Kinvoa arms planning experiments with 5, 10, 15 obstacles under 1.0s time limit $\downarrow$}
        \label{tab:multiarm_planningtime_1.0s}
\end{table*}

Tables \ref{tab:multiarm_planningtime_1.0s} and \ref{tab:multiarm_consevaltime_1.0s} present similar results for robots with $14$ and $21$ degrees of freedom.
Similar to the previous results, \methodname computes the constraints and constraint gradients almost twice as fast as \armtd.
Furthermore, compared to \methodname, \armtd struggles to generate motion plans in the allotted time.


\subsection{Motion Planning Experiments}\label{subsec:baselines}
We evaluate the performance of \methodname on on two sets of scenarios.
The first set of scenes, Random Obstacles, shows that \methodname can generate safe trajectories for arbitrary tasks.
For the $7$ DOF arm, the scene contains $n_{\mathcal{O}}=10$, $20$, or $40$ axis-aligned boxes that are $20$cm on each side.
For the $14$ and $21$ DOF arms, the scene contains $n_{\mathcal{O}}=5$, $10$, or $15$ obstacles of the same size.
For each number of obstacles, we generate 100 tasks with obstacles placed randomly such that the start and goal configurations are collision-free.
Note that a solution to the goal may not be guaranteed to exist.
The second set of tasks, Hard Scenarios, contains 14 challenging tasks with the number of obstacles varying from $5$ to $20$.
We compare the performance of \methodname to \armtd, \chomp, \trajopt, and \mpot on both sets of scenarios for the $7$ DOF arm.
We compare the performance of \methodname to \armtd on Random Obstacles for the $14$ and $21$ DOF robots.

In the tests with a $7$ DOF arm, \methodname and \armtd are given a planning time limit of $0.5$s and a maximum of 150 planning iterations to reach the goal. For the tests with $14$ and $21$ DOF arms, \methodname and \armtd are given a planning time limit of $1.0$s and a maximum of 150 planning iterations. 
A failure occurs if \methodname or \armtd fails to find a plan for two consecutive planning iterations or if a collision occurs.
Note the meshes for the robot and obstacles are used to perform ground truth collision checking within the simulator.
Finally, \methodname and \armtd are given a straight-line high-level planner for both the Random Obstacles scenarios and the Hard Scenarios.

Table \ref{tab:random_obstacles_0.5s} presents the results for the Random Obstacles scenarios where \methodname and \armtd are given a maximum planning time limit $0.5$s.
\methodname ($\pi/6$) achieves the highest success rate across all obstacles followed by \methodname $(\pi/24)$. 
Notably, the performance of \armtd degrades drastically as the number of obstacles is increased.
This is consistent with \armtd's slower constraint evaluation presented in Table \ref{tab:consevalruntime3d7link0.5s}, \armtd's mean planning time exceeding the limit, and \armtd's overly conservative reachable sets shown in Fig \ref{fig:reachset_comparison}.
Although \armtd remains collision-free, these results suggest that the optimization problem is more challenging for \armtd to solve.
In contrast \chomp, \trajopt, and \mpot exhibit much lower success rates.
Note that \trajopt and \mpot frequently crash whereas \chomp does not return feasible trajectories.

Table \ref{tab:multiarm_success1.0} compares the success rate of \methodname to \armtd on Random Obstacles as the degrees of freedom of the robot are varied.
For $14$ degrees of freedom and $n_{\mathcal{O}} = 5$ obstacles, \armtd$(\pi/24)$ has more successes though \mbox{\methodname$(\pi/24)$} and \methodname$(\pi/6)$ remain competitive.
Note, in particular, that \methodname $(\pi/6)$'s success rate is more than double that of \methodname$(\pi/6)$ for $10$ and $15$ obstacles.
For $21$ degrees of freedom, the performance gap between \methodname and \armtd is much larger as \armtd fails to solve any task in the presence of $10$ and $15$ obstacles.

Table \ref{tab:hard_scenarios_0.5s} presents results for the Hard Scenarios. 
\mpot outperforms all of the other methods with 6 successes and 8 collisions, followed by \methodname$(\pi/6)$ which obtains 5 successes. 
Note, however, that \methodname remains collision-free in all scenarios.


\begin{table}[t]
    \centering
    \begin{tabular}{|c||c||c||c||}
    \hline
    Methods & \multicolumn{3}{c||}{\# Successes}\\\hline
    \# Obstacles & 10 & 20 & 40 \\
    \hline \hline
    \methodname{} $(\pi/24)$ & 79 & 61 & 34 \\\hline
    \armtd $(\pi/24)$ & 79 & 50 & 2 \\\hline
    \methodname{} $(\pi/6)$   & \textbf{87} & \textbf{62} & \textbf{40} \\\hline
    \armtd $(\pi/6)$ & 56 & 17 & 0 \\\hline
    \chomp \cite{Zucker2013chomp} & 30  & 9  & 4 \\
    \hline
    \trajopt \cite{Schulman2014trajopt} & 33 (\textcolor{red}{67}) & 9 (\textcolor{red}{91}) & 6 (\textcolor{red}{94}) \\
    \hline
    \mpot \cite{le2023mpot} & 57 (\textcolor{red}{43}) & 22 (\textcolor{red}{78}) & 9 (\textcolor{red}{91}) \\\hline
    \end{tabular}
    \caption{Number of successes for \methodname{}, \armtd, \mpot, \chomp, and Trajopt in Kinova planning experiment with 10, 20, and 40 \textbf{Random Obstacles} $\uparrow$. \textcolor{red}{Red} indicates the number of failures due to collision.}
    \label{tab:random_obstacles_0.5s}
\end{table}

\begin{table}[t]
        \centering
            \begin{tabular}{ | c || c || c || c || c || c || c ||}
                \hline 
                Methods & \multicolumn{6}{c||}{\# Successes}\\\hline
                \# DOF & \multicolumn{3}{c||}{14} & \multicolumn{3}{c||}{21} \\  \hline
                \# Obstacles & 5 & 10 & 15 & 5 & 10 & 15 \\  \hline \hline
                \methodname{} $(\pi/24)$ & 92 & 79 & 62 & \textbf{73} & \textbf{46} & \textbf{29} \\\hline
                \armtd $(\pi/24)$ & \textbf{96} & 75 & 34 & \textbf{73} & 0 & 0 \\\hline
                \methodname{} $(\pi/6)$   & 94 & \textbf{83}  & \textbf{68} & 67 & 36 & 13 \\ \hline
                \armtd $(\pi/6)$          & 83 & 39 & 6 & 41 & 0 & 0  \\ \hline
                \end{tabular}
        \caption{Number of successes for \methodname{} and \armtd in 2 and 3 Kinvoa arms planning experiment with 5, 10, 15 obstacles under 1.0s time limit $\uparrow$}
        \label{tab:multiarm_success1.0}
\end{table}

\begin{table}[t]
    \centering
    \begin{tabular}{|c||c||c||c||c||}
    \hline
    Methods & \# Successes\\\hline
    \hline \hline
    \methodname{} $(\pi/24)$   & 3 \\\hline
    \armtd $(\pi/24)$ & 3 \\\hline
    \methodname{} $(\pi/6)$ & 5 \\\hline
    \armtd $(\pi/6)$ & 4 \\\hline
    \chomp \cite{Zucker2013chomp} & 2 \\
    \hline
    \trajopt \cite{Schulman2014trajopt} & 0 (\textcolor{red}{14})  \\
    \hline
    \mpot \cite{le2023mpot} & \textbf{6} (\textcolor{red}{8}) \\\hline
    \end{tabular}
    \caption{Number of success for \methodname{}, \armtd, \mpot, \chomp, and \trajopt in Kinova planning experiment on 14 \textbf{Hard Scenarios} $\uparrow$. \textcolor{red}{Red} indicates the number of failures due to collision.}\label{tab:hard_scenarios_0.5s}
\end{table}

\section{Conclusion}
We present \methodname, a method to perform safe, real-time manipulator motion planning in cluttered environments.
We introduce the \SFO to overapproximate the reachable set of a serial manipulator robot in a manner that is less conservative than previous approaches.
We also introduce an algorithm to compute the exact signed distance between a point and a 3D zonotope.
We use the spherical forward occupancy in conjunction with the exact signed distance algorithm to compute obstacle-avoidance constraints in a trajectory optimization problem.
By leveraging reachability analysis, we demonstrate that \methodname is certifiably-safe, yet less conservative than previous methods.
Furthermore, we demonstrate \methodname outperforming various state-of-the-art methods on challenging motion planning tasks.
Despite \methodname' performance it is inherently limited if it is required to reason about numerous individual objects while generating motion plans.
The major downside of our approach is that it requires the enumeration the vertices of 3D zonotopes prior to planning.
However, we believe this provides a unique opportunity to combine rigorous model-based methods with deep learning-based methods for probabilistically-safe robot motion planning.
Therefore, future directions aim to incorporate neural implicit scene representations into \methodname while accounting for uncertainty.





\renewcommand{\bibfont}{\normalfont\footnotesize}
{\renewcommand{\markboth}[2]{}
\printbibliography}

\appendices
\clearpage
\section{Spherical Forward Occupancy Proof}\label{appendix:sfo_proof}
\begin{figure*}[ht]
    \centering
    \includegraphics[width=\textwidth]{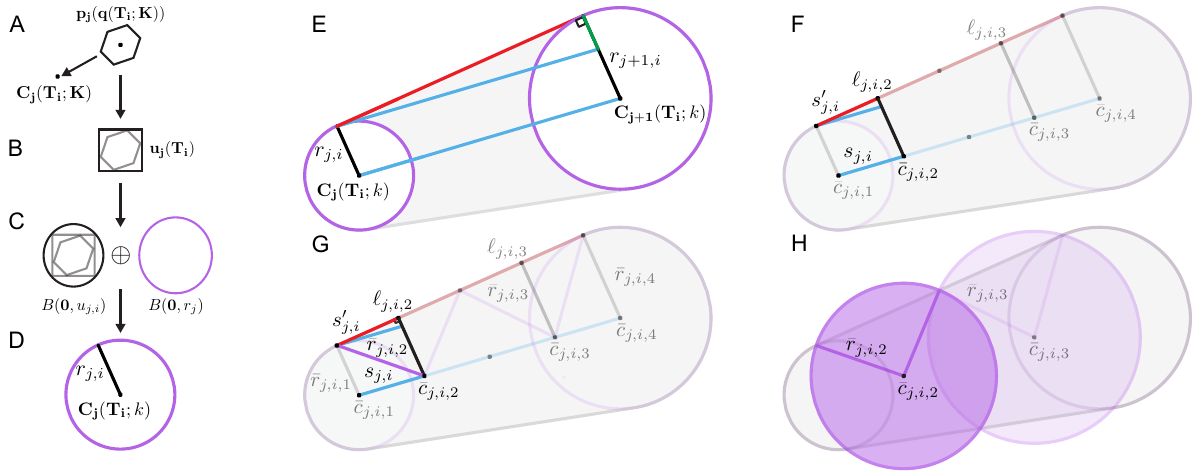}
    \caption{ Illustration of the \SJO and \SFO construction proofs in 2D.}
    \label{fig:sfo_2D}
\end{figure*}

This section provides the proofs for Lemma \ref{lem:sjo} and Theorem \ref{thm:sfo} which are visually illustrated in Fig. \ref{fig:sfo_2D}.

\subsection{Proof of Lemma \ref{lem:sjo}}\label{appendix:sjo}
\begin{proof}
Using Lem. \ref{lem:PZFK} and Alg. \ref{alg:compose_fk} one can compute $\pz{p_j}(\pzqi)$, which overapproximates the position of the $j$\ts{th} joint in the workspace.
We then split $\pz{p_j}(\pzqi)$ (Fig. \hyperref[fig:sfo_2D]{6A}) into
\begin{align}
    \pz{p_j}(\pzqi) &= \PZ{c, \pzgi, \pzei, \pzv, h_j, y_j} \\ 
    &= \PZ{\pzgi, \pzei, \pzv, \mathbf{0}, \mathbf{0}}  
    \oplus \PZ{\mathbf{0}, \mathbf{0}, \mathbf{0}, h_j, y_j}  \\
    &= \pzCjiK \oplus \pzuj,
\end{align}
where $\pzCjiK$ is a polynomial zonotope whose generators that are only dependent on $\pz{K}$ and $\pzuj$ is a zonotope that is independent of $\pz{K}$.
This means slicing $\pzCjiK$ at a particular trajectory parameter $k \in K$ yields a single point given by $\pzCjik$.

We next overapproximate $\pz{p_j}(\pzqi)$.
We start by defining the vector $u_{j,i,l}$ such that
\begin{equation}
    u_{j,i,l} = \setop{sup}(\pzuj \odot \hat{e}_l) \quad \forall l \in \{1,2,3\},
\end{equation}
where $\hat{e}_l$ is a unit vector in the standard orthogonal basis.
Together, these vectors $\{u_{j,i,l}\}_{l=1}^{3}$ create an axis-aligned cube that bounds $\pzuj$ (Fig. \hyperref[fig:sfo_2D]{6B}).
Using the L2 norm of $\sum_{l=1}^{3} u_{j,i,l}$ allows one to then bound $\pz{p_j}(\pzqi)$ (Fig. \hyperref[fig:sfo_2D]{6C}, left) as
\begin{align}
    \pz{p_j}(\pzqi) \subset \pzCjiK \oplus B(\mathbf{0}, u_{j,i}) ,
\end{align}
where
\begin{equation}
    u_{j,i} = \|u_{j,i,1} + u_{j,i,2} + u_{j,i,3}\|_2.
\end{equation}
Finally, letting $r_j$ denote the sphere radius defined in Assum. \ref{assum:joint_link_occupancy} (Fig. \hyperref[fig:sfo_2D]{6C}, right), we overapproximate the volume occupied by the $j$\ts{th} joint as

\begin{align}
    \SjiK &= \pzCjiK \oplus B(\mathbf{0}, u_{j,i}) \oplus B(\mathbf{0}, r_j) \nonumber \\
    &=\pzCjiK \oplus B(\mathbf{0}, r_{j,i}),
\end{align}
where $r_{j,i} = r_j + u_{j,i}$.

Then for all $i \in \Nt$ and $j \in \Nq$ we define the \SJO{} ($\sjo$) as the collection of sets $\SjiK$:
\begin{equation}
    \sjo = \left\{\SjiK \, : \, \forall i\in \Nt, j\in \Nq \right\}.
\end{equation}
\noindent Note that for any $k \in K$ we can slice each element of $\sjo$ to get a sphere centered at $\pzCjik$ with radius $r_{j,i}$ and the \emph{sliced} \SJO (Fig. \hyperref[fig:sfo_2D]{6D}):
\begin{equation}
    \Sjik = B(\pzCjik, r_{j,i}).
\end{equation}
\begin{equation}
    \sjo(k) = \left\{\Sjik \, : \, \forall i\in \Nt, j\in \Nq \right\}.
\end{equation}
\end{proof}

\subsection{Proof of Theorem \ref{thm:sfo}}\label{appendix:sfo}

\begin{proof}\label{proof:sfo}
Let $\pzCjik$ and $r_{j,i}$ be the center and radius of $\Sjik$, respectively.
We begin by dividing the line segments connecting the centers and tangent points of $\Sjik$ and $\Sjpik$ into $2(\ns-2)$ equal line segments of length
\begin{equation}\label{eq:sphere_center_line_segment}
    \sjik = \frac{\left\|\pzCjpik - \pzCjik\right\|}{2(\ns-2)}
\end{equation}
and
\begin{align}
    \spjik &= \biggl(  \frac{\left\|\pzCjpik - \pzCjik\right\|}{2(\ns-2)} - (\frac{r_{j+1,i} - r_{j,i}}{2(\ns-2)}) \biggr)^\frac{1}{2} \nonumber \\
    &=  \biggl(  \sjik^2 - \bigl( \frac{r_{j+1,i} - r_{j,i}}{2(\ns-2)} \bigr)^2 \biggr)^\frac{1}{2}
    \label{eq:sphere_tangent_line_segment} 
\end{align}
respectively.
By observing that the tangent line is perpendicular to both spheres and then applying the Pythagorean Theorem  to the right triangle with leg (Fig. \hyperref[fig:sfo_2D]{6E}, green) and hypotenuse (Fig. \hyperref[fig:sfo_2D]{6E}, light blue) lengths given by $(r_{j+1,i} - r_{j,i})$ and $2 \cdot (\ns-2) \cdot \sjik$, respectively, one can obtain \eqref{eq:sphere_tangent_line_segment}.

We then choose $\ns$ sphere centers (Fig. \hyperref[fig:sfo_2D]{6F}) such that
\begin{equation}
    \bar{c}_{j,i,1}(k) = \pzCjik,
\end{equation}
\begin{equation}
    \bar{c}_{j,i,\ns}(k) = \pzCjpik,
\end{equation}
and
\begin{equation}
    \bar{c}_{j,i,m+1}(k) = \pzCjik + \frac{2m - 1}{2(\ns-2)} \left(\pzCjpik - \pzCjik\right),
\end{equation}
where $1 \leq m\leq \ns-2$.

Next, let 
\begin{equation}
    \bar{r}_{j,i,1}(k) = r_{j,i}
\end{equation}
and 
\begin{equation}
    \bar{r}_{j,i,\ns}(k) = r_{j+1,i}.
\end{equation}
Before constructing the remaining radii, we first compute $\ell_{j,i,m+1}(k)$, which is the length of the line segment (Fig. \hyperref[fig:sfo_2D]{6F}, black) that extends perpendicularly from the tapered capsule surface through center $\bar{c}_{j,i,m+1}(k)$:
\begin{equation}
    \ell_{j,i,m+1}(k) = r_{j,i} + \frac{2m - 1}{2(\ns-2)} \left(r_{j+1,i} - r_{j,i}\right),
\end{equation}
where $1 \leq m\leq \ns-2$.
\noindent
We construct the remaining radii such that adjacent spheres intersect at a point on the tapered capsule's tapered surface. 
This ensures each sphere extends past the surface of the tapered capsule to provide an overapproximation.
Then, $\bar{r}_{j,i,m+1}(k)$ is obtained by applying the Pythagorean Theorem to the triangle with legs $\ell_{j,i,m+1}(k)$ (Fig. \hyperref[fig:sfo_2D]{6G}, black) and $\spjik$ (Fig. \hyperref[fig:sfo_2D]{6G}, red), respectively:
\begin{align}
    \bar{r}_{j,i,m+1}(k)  &= \left(\ell_{j,i,m+1}^2 + s'_{j,i}(k)^2\right)^{\frac{1}{2}} \nonumber \\
    &= \left(\ell_{j,i,m+1}^2 + \sjik^2 - \bigl( \frac{r_{j+1,i} - r_{j,i}}{2(\ns-2)} \bigr)^2 \right)^{\frac{1}{2}}
\end{align}

Thus, for a given $k\in K$, we define $\sfok$ (Fig. \hyperref[fig:sfo_2D]{6H}, purple) to be the set
\begin{equation}
    \sfojk = \left\{\bar{S}_{j,i,m}(\pzqki) \, : \, 1 \leq m \leq \ns \right\},
\end{equation}
where
\begin{equation}
    \bar{S}_{j,i,m}(\pzqki) = B\left(\bar{c}_{j,i,m}(k), \bar{r}_{j,i,m}(k)\right).
\end{equation}
By construction, the tapered capsule $\TCjik$ is a subset of $\sfojk$ for each  $j \in \Nq$,  $k \in \pz{K}$, and $t \in \pz{T_i}$.
Further note that the construction of $\sfojk$ is presented for the 2D case.
However, this construction remains valid in 3D if one restricts the analysis to any two dimensional plane that intersects the center of each $\pzCjik$ and $\pzCjpik$ .
\end{proof}

\begin{figure*}[t]
    \centering
    \includegraphics[width=\textwidth]{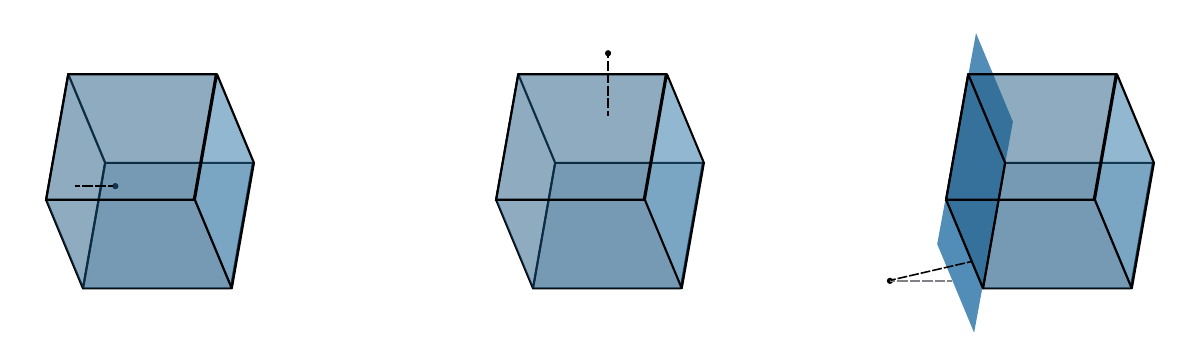}
    \caption{ Illustration of signed distance between a point and a zonotope in 3D. The first panel shows a point inside the zonotope. The second panel shows a point outside the zonotope, but whose projection lies on a single face of the zonotope. The third panel shows a point outside the zonotope, but whose projection lies on an edge of the zonotope.
    }
    \label{fig:projection_proof}
\end{figure*}
\subsection{Proof of Lemma \ref{lem:sdf}}\label{appendix:sdf}
There are three cases to consider.
In the first case, $c$ is inside the zonotope $\mathcal{O}$ (Fig. \ref{fig:projection_proof}, left). 
Let $(A,b)$ be the polytope representation of $\mathcal{O}$ such that $A \in R^{m \times n}$ and $b \in \R^n$.
Then each row of $Ac-b$ corresponds to the signed distance to each hyperplane comprising the faces of $\mathcal{O}$.
Because $c \in \mathcal{O}$, each row of $Ac-b$ is less than or equal to zero.
Therefore, the signed distance to $\mathcal{O}$ is given by
\begin{equation}
    \sdf(c; \mathcal{O}) = \max(Ac-b).
\end{equation}

In the second case, $c$ is not inside $\mathcal{O}$ but there exists $i_A \in [1,m]$ such that
\begin{align}
    p_{\text{face}} &= \proj(c; \mathcal{H}_{i_A}) \\ 
    &=  c - (Ac-b)_{i_A} \cdot A_{i_A}^{T} \in \mathcal{O},
\end{align}
where $p_{\text{face}}$ is the projection of $c$ onto the $i_A$\ts{th} hyperplane $\mathcal{H}_{i_A}$ of $\mathcal{O}$.
Then the signed distance to $\mathcal{O}$ is given by
\begin{equation}
    \sdf(c; \mathcal{O}) = (Ac-b)_{i_A}.
\end{equation}

In the final case, $c$ is not inside $\mathcal{O}$, but the smallest positive signed distance to each hyperplane of $\mathcal{O}$ (Fig. \ref{fig:projection_proof}, right) lies on an edge of $\mathcal{O}$. 
Then the signed distance to $\mathcal{O}$ is given by
\begin{equation}
    \sdf(c; \mathcal{O}) \underset{E_{i_E} \in E}{\min} \dist(c; E_{i_E}).
\end{equation}


\section{Additional Results}\label{appendix:tables}
\begin{table}[ht]
        \centering
            \begin{tabular}{ | c || c | c | c ||}
                \hline 
                Methods & \multicolumn{3}{c||}{mean constraint evaluation time [ms]} \\\hline
                \# Obstacles (s) & 10 & 20 & 40 \\\hline \hline
                \methodname{} $(\pi/24)$  & \textbf{3.1 ± 0.1} & \textbf{3.8 ± 0.1} & \textbf{5.1 ± 0.1}\\ \hline
                ARMTD $(\pi/24)$          & 4.1 ± 0.3 & 5.4 ± 0.5 & 8.1 ± 0.6 \\ \hline
                \methodname{} $(\pi/6)$   & \textbf{3.1 ± 0.1} & \textbf{3.8 ± 0.1} & \textbf{5.1 ± 0.1} \\ \hline
                ARMTD $(\pi/6)$ & 4.1 ± 0.3 & 5.4 ± 0.4 & 7.6 ± 0.3  \\ \hline
                \end{tabular}
        \caption{Mean runtime for constraint and constraint gradient evaluation for \methodname{} and \armtd in Kinova planning experiments with 10, 20, and 40 obstacles under 0.25s time limit $\downarrow$}
        \label{tab:consevalruntime3d7link0.25s}
\end{table}
\begin{table}[ht]
        \centering
            \begin{tabular}{ | c || c | c | c ||}
                \hline 
                Methods & \multicolumn{3}{c||}{mean constraint evaluation time [ms]} \\\hline
                \# Obstacles (s) & 10 & 20 & 40 \\\hline \hline
                \methodname{} $(\pi/24)$  & \textbf{3.1 ± 0.1} & \textbf{3.8 ± 0.1} & \textbf{5.1 ± 0.1}\\ \hline
                ARMTD $(\pi/24)$  & 4.1 ± 0.3 & 5.3 ± 0.1 & 7.8 ± 0.4 \\ \hline
                \methodname{} $(\pi/6)$   & \textbf{3.1 ± 0.1} & \textbf{3.8 ± 0.1} & \textbf{5.1 ± 0.1} \\ \hline
                ARMTD $(\pi/6)$ & 4.0 ± 0.2 & 5.2 ± 0.2 & 8.8 ± 0.7 \\ \hline
                \end{tabular}
        \caption{Mean runtime for constraint and constraint gradient evaluation for \methodname{} and \armtd in Kinova planning experiments with 10, 20, and 40 obstacles under 0.15s time limit $\downarrow$}
        \label{tab:consevalruntime3d7link0.15s}
\end{table}
\begin{table}[ht]
        \centering
            \begin{tabular}{ | c || c | c | c ||}
                \hline 
                Methods & \multicolumn{3}{c||}{mean planning time [s]} \\\hline
                \# Obstacles (s) & 10 & 20 & 40 \\\hline \hline
                \methodname{} $(\pi/24)$  & \textbf{0.12 ± 0.04} & \textbf{0.14 ± 0.05} & \textbf{0.17 ± 0.05}\\ \hline
                ARMTD $(\pi/24)$          & 0.16 ± 0.04 & 0.24 ± 0.03 & \textcolor{red}{0.27 ± 0.02} \\ \hline
                \methodname{} $(\pi/6)$   & 0.14 ± 0.04 & 0.16 ± 0.04 & 0.20 ± 0.04 \\ \hline
                ARMTD $(\pi/6)$           & 0.20 ± 0.04 & \textcolor{red}{0.26 ± 0.02} & \textcolor{red}{0.27 ± 0.02} \\ \hline
                \end{tabular}
        \caption{Mean per-step planning time for \methodname{} and ARMTD in Kinova planning experiment with 10, 20, and 40 obstacles under 0.25s time limit $\downarrow$. \textcolor{red}{Red} indicates that the average planning time limit has been exceeded.}
        \label{tab:planningtime3d7link0.25s}
\end{table}

\begin{table}[ht]
        \centering
            \begin{tabular}{ | c || c | c | c ||}
                \hline 
                Methods & \multicolumn{3}{c||}{mean planning time [s]} \\\hline
                \# Obstacles (s)  & 10 & 20 & 40 \\\hline \hline
                \methodname{} $(\pi/24)$   & \textbf{0.11 ± 0.02} & \textbf{0.13 ± 0.02} & \textbf{0.14 ± 0.02}\\ \hline
                ARMTD $(\pi/24)$           & 0.14 ± 0.02 & \textcolor{red}{0.16 ± 0.03} & \textcolor{red}{0.17 ± 0.03} \\ \hline
                \methodname{} $(\pi/6)$    & 0.13 ± 0.02 & 0.14 ± 0.02 & 0.15 ± 0.02 \\ \hline
                ARMTD $(\pi/6)$ & \textcolor{red}{0.16 ± 0.02} & \textcolor{red}{0.16 ± 0.03} & \textcolor{red}{0.18 ± 0.03} \\ \hline
                \end{tabular}
        \caption{Mean per-step planning time for \methodname{} and ARMTD in Kinova planning experiment with 10, 20, and 40 obstacles under 0.15s time limit $\downarrow$. \textcolor{red}{Red} indicates that the average planning time limit has been exceeded.}
        \label{tab:planningtime3d7link0.15s}
\end{table}

\begin{table}[t]
        \centering
            \begin{tabular}{ | c || c || c || c || c || c || c ||}
                \hline 
                Methods & \multicolumn{6}{c||}{\# Successes}\\\hline
                \# DOF & \multicolumn{3}{c||}{14} & \multicolumn{3}{c||}{21} \\  \hline
                \# Obstacles & 5 & 10 & 15 & 5 & 10 & 15 \\  \hline \hline
                \methodname{} $(\pi/24)$ & \textbf{74} & \textbf{47} &  \textbf{27}  & 0  & 0 & 0 \\\hline
                \armtd $(\pi/24)$ & 70 & 3  & 0 & 0 & 0 & 0 \\\hline
                \methodname{} $(\pi/6)$ & 55  & 42  & 12  & 0  & 0 & 0 \\ \hline
                \armtd $(\pi/6)$ & 35 & 0 & 0 & 0 & 0  & 0  \\ \hline
                \end{tabular}
        \caption{Number of successes for \methodname{} and \armtd in 2 and 3 Kinvoa arms planning experiment with 5, 10, 15 obstacles under 0.5s time limit $\uparrow$ }
        \label{tab:multiarm_success0.5}
\end{table}


\begin{table*}[t]
        \centering
            \begin{tabular}{ | c || c | c | c || c | c  | c ||}
                \hline 
                Methods & \multicolumn{6}{c||}{mean constraint evaluation time [ms]}\\\hline
                \# DOF & \multicolumn{3}{c||}{14} & \multicolumn{3}{c||}{21} \\  \hline
                \# Obstacles & 5 & 10 & 15 & 5 & 10 & 15 \\ \hline \hline
                \methodname{} $(\pi/24)$ & \textbf{3.9 ± 0.2} & \textbf{4.7 ± 0.1} & \textbf{5.4 ± 0.2} & 4.9 ± 0.2 & \textbf{6.0 ± 0.1} & \textbf{7.2 ± 0.1} \\ \hline
                \armtd $(\pi/24)$ & 8.4 ± 0.6 & 9.3 ± 0.6 & 10.6 ± 0.4 & 12.6 ± 0.6 & 14.3 ± 0.5 & 17.1 ± 1.3 \\ \hline
                \methodname{} $(\pi/6)$ & \textbf{3.9 ± 0.1} & \textbf{4.6 ± 0.1} & \textbf{5.4 ± 0.1} & \textbf{4.8 ± 0.1} & \textbf{6.0 ± 0.2} & \textbf{7.2 ± 0.2} \\ \hline
                \armtd $(\pi/6)$ & 8.3 ± 0.6 & 9.0 ± 0.2 & 10.6 ± 0.1 & 13.0 ± 0.7 & 14.6 ± 0.8 & 16.4 ± 0.9 \\ \hline
                \end{tabular}
        \caption{Mean runtime for a constraint and constraint gradient evaluation for \methodname{} and \armtd in 2 and 3 Kinvoa arms planning experiment with 5, 10, 15 obstacles under 0.5s time limit $\downarrow$}
        \label{tab:multiarm_consevaltime_0.5s}
\end{table*}

\begin{table*}[t]
        \centering
            \begin{tabular}{ | c || c | c | c || c | c | c | c ||}
                \hline 
                Methods & \multicolumn{6}{c||}{mean planning time [s]}\\\hline
                \# DOF & \multicolumn{3}{c||}{14} & \multicolumn{3}{c||}{21} \\  \hline
                \# Obstacles & 5 & 10 & 15 & 5 & 10 & 15 \\ \hline \hline
                \methodname{} $(\pi/24)$ & \textbf{0.32 ± 0.06} & \textbf{0.35 ± 0.07} & \textbf{0.37 ± 0.08} & \textcolor{red}{0.51 ± 0.02} & \textcolor{red}{0.51 ± 0.01} & \textcolor{red}{0.52 ± 0.02} \\ \hline
                \armtd $(\pi/24)$ & 0.37 ± 0.06 & 0.49 ± 0.04 & \textcolor{red}{0.52 ± 0.02} & \textcolor{red}{0.54 ± 0.02} & \textcolor{red}{0.55 ± 0.03} & \textcolor{red}{0.55 ± 0.03} \\ \hline
                \methodname{} $(\pi/6)$ & 0.37 ± 0.08 & 0.40 ± 0.08 & 0.44 ± 0.07 & \textcolor{red}{0.53 ± 0.02} & \textcolor{red}{0.52 ± 0.01} & \textcolor{red}{0.53 ± 0.01} \\ \hline
                \armtd $(\pi/6)$ & 0.42 ± 0.07 & \textcolor{red}{0.52 ± 0.02} & \textcolor{red}{0.54 ± 0.14} & \textcolor{red}{0.53 ± 0.02} & \textcolor{red}{0.56 ± 0.02} & \textcolor{red}{0.54 ± 0.02} \\ \hline
                \end{tabular}
         \caption{Mean per-step planning time for \methodname{} and \armtd in 2 and 3 Kinvoa arms planning experiment with 5, 10, 15 obstacles under 0.5s time limit $\downarrow$}
        \label{tab:multiarm_planningtime_0.5s}
        \vspace{13cm}
\end{table*}

\end{document}